\documentclass{article}
\usepackage{setspace}
\usepackage{cite}
\usepackage[pdftex]{graphicx}

\usepackage[cmex10]{amsmath}
\usepackage[tight,footnotesize]{subfigure}
\usepackage{url}

\usepackage{amssymb}
\usepackage{amsthm}
\usepackage{mdwlist} 
\usepackage{color}

\DeclareMathOperator*{\argmin}{argmin}
\newcommand{\ol}[1]{\overline{#1}}

\newcommand{\dualpar}{a}
\newcommand{\hyperplane}{\bm{w}}

\newcommand{\hypothesis}{h}
\newcommand{\rankfunc}{f}
\newcommand{\funspace}{\mathcal{H}}

\newcommand{\kernelf}{K} 

\newcommand{\lossfunction}{\mathcal{L}}

\newcommand{\regparam}{\lambda}
\newcommand{\nodeset}{\mathcal{V}}
\newcommand{\edgeset}{\mathcal{E}}

\newcommand{\weightfunc}{Q}
\newcommand{\edgeweight}{y}

\newcommand{\node}{v}
\newcommand{\edge}{e}
\newcommand{\nodecount}{p}
\newcommand{\edgecount}{q}
\newcommand{\feaspacecount}{r}

\newcommand{\mbr}{\mathbb{R}}

\newcommand{\bm}[1]{\mathbf{#1}}

\newcommand{\anymatrix}{\bm{M}}
\newcommand{\othermatrix}{\bm{N}}

\newtheorem{theorem}{Theorem}[section]

\newtheorem{proposition}[theorem]{Proposition}

\newtheorem{definition}[theorem]{Definition}

\begin{document}
%
\title{A kernel-based framework for learning \\ graded relations from data}

\author{Willem Waegeman,
        Tapio Pahikkala,
        Antti Airola,\\
        Tapio Salakoski,
        Michiel Stock,
        Bernard De Baets} 

\date{}

\maketitle

\begin{abstract}
Driven by a large number of potential applications in areas like bioinformatics, information retrieval and social network analysis, the problem setting of inferring relations between pairs of data objects has recently been investigated quite intensively in the machine learning community. To this end, current approaches typically consider datasets containing crisp relations, so that standard classification methods can be adopted. However, relations between objects like similarities and preferences are often expressed in a graded manner in real-world applications. A general kernel-based framework for learning relations from data is introduced here. It extends existing approaches because both crisp and graded relations are considered, and it unifies existing approaches because different types of graded relations can be modeled, including symmetric and reciprocal relations. This framework establishes important links between recent developments in fuzzy set theory and machine learning. Its usefulness is demonstrated through various experiments on synthetic and real-world data.
\end{abstract}


\section{Introduction}
Relational data occurs in many predictive modeling tasks, such as forecasting the winner in two-player computer games \cite{Bowling2006}, predicting proteins that interact with other proteins in bioinformatics \cite{Yamanishi2004}, retrieving documents that are similar to a target document in text mining \cite{Yang2009}, investigating the persons that are friends of each other on social network sites \cite{Taskar2004}, etc. All these examples represent fields of application in which specific machine learning and data mining algorithms have been successfully developed to infer relations from data; pairwise relations, to be more specific.

The typical learning scenario in such situations can be summarized as follows. Given a dataset of known relations between pairs of objects and a feature representation of these objects in terms of variables that might characterize the relations, the goal usually consists of inferring a statistical model that takes two objects as input and predicts whether the relation of interest occurs for these two objects. Moreover, since one aims to discover unknown relations, a good learning algorithm should be able to construct a predictive model that can generalize for unseen data, i.e., pairs of objects for which at least one of the two objects was not used to construct the model. As a result of the transition from predictive models for single objects to pairs of objects, new advanced learning algorithms need to be developed, resulting in new challenges with regard to model construction, computational tractability and model assessment.

As relations between objects can be observed in many different forms, this general problem setting provides links to several subfields of machine learning, like statistical relational learning \cite{Deraedt2009}, graph mining \cite{Vert2005}, metric learning \cite{Xing2002} and preference learning \cite{Hullermeier2010a}. More specifically, from a graph-theoretic perspective, learning a relation can be formulated as learning edges in a graph where the nodes represent information of the data objects; from a metric learning perspective, the relation that we aim to learn should satisfy some well-defined properties like positive definiteness, transitivity or the triangle inequality; and from a preference learning perspective, the relation expresses a (degree of) preference in a pairwise comparison of data objects.

The topic of learning relations between objects is also closely related to recent developments in fuzzy set theory. This article will elaborate on these connections via two important contributions: (1) the extension of the typical setting of learning crisp relations to real-valued and ordinal-valued relations and (2) the inclusion of domain knowledge about relations into the inference process by explicit modeling of mathematical properties of these relations. For algorithmic simplicity, one can observe that many approaches only learn crisp relations, that is relations with only 0 and 1 as possible values, so that standard binary classifiers can be modified. In this context, consider examples as inferring protein-protein interaction networks or metabolic networks in bioinformatics \cite{Yamanishi2004,Geurts2007}.

However, graded relations are observed in many real-world applications \cite{Doignon1986}, resulting in a need for new algorithms that take graded relational information into account. Furthermore, the properties of graded relations have been investigated intensively in the recent fuzzy logic literature\footnote{Often the term fuzzy relation is used in the fuzzy set literature to refer to graded relations. However, fuzzy relations should be seen as a subclass of graded relations. For example, reciprocal relations should not be considered as fuzzy relations, because they often exhibit a probabilistic semantics rather than a fuzzy semantics.}, and these properties are very useful to analyze and improve current algorithms. Using mathematical properties of graded relations, constraints can be imposed for incorporating domain knowledge in the learning process, to improve predictive performance or simply to guarantee that a relation with the right properties is learned. This is definitely the case for properties like transitivity when learning similarity relations and preference relations -- see e.g.\ \cite{Switalski2000,DeBaets2005,DeBaets2006,Diaz2007}, but even very basic properties like symmetry, antisymmetry or reciprocity already provide domain knowledge that can steer the learning process. For example, in social network analysis, the notion ``person A being a friend of person B" should be considered as a symmetric relation, while the notion ``person A defeats person B in a chess game" will be antisymmetric (or, equivalently, reciprocal). Nevertheless, many examples exist, too, where neither symmetry nor antisymmetry necessarily hold, like the notion ``person A trusts person B".

\begin{figure}
\begin{minipage}[c]{0.38\textwidth}
\begin{center}
\includegraphics[scale=0.3]{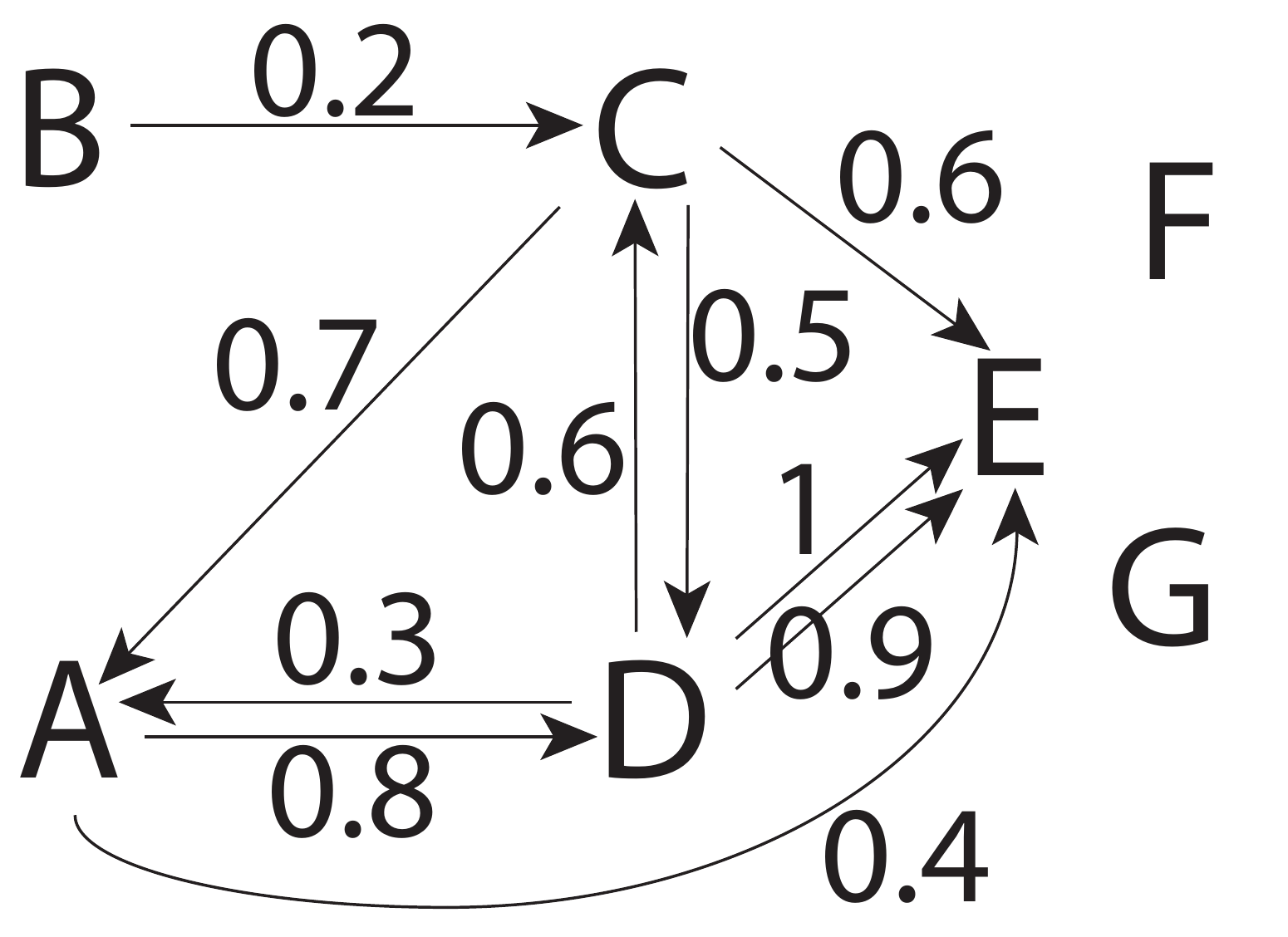}
\end{center}
\end{minipage}
\begin{minipage}[c]{0.58\textwidth}
\centering
\subfigure[C, R, T]{
\label{fig:gr:000} 
\includegraphics[scale=0.19]{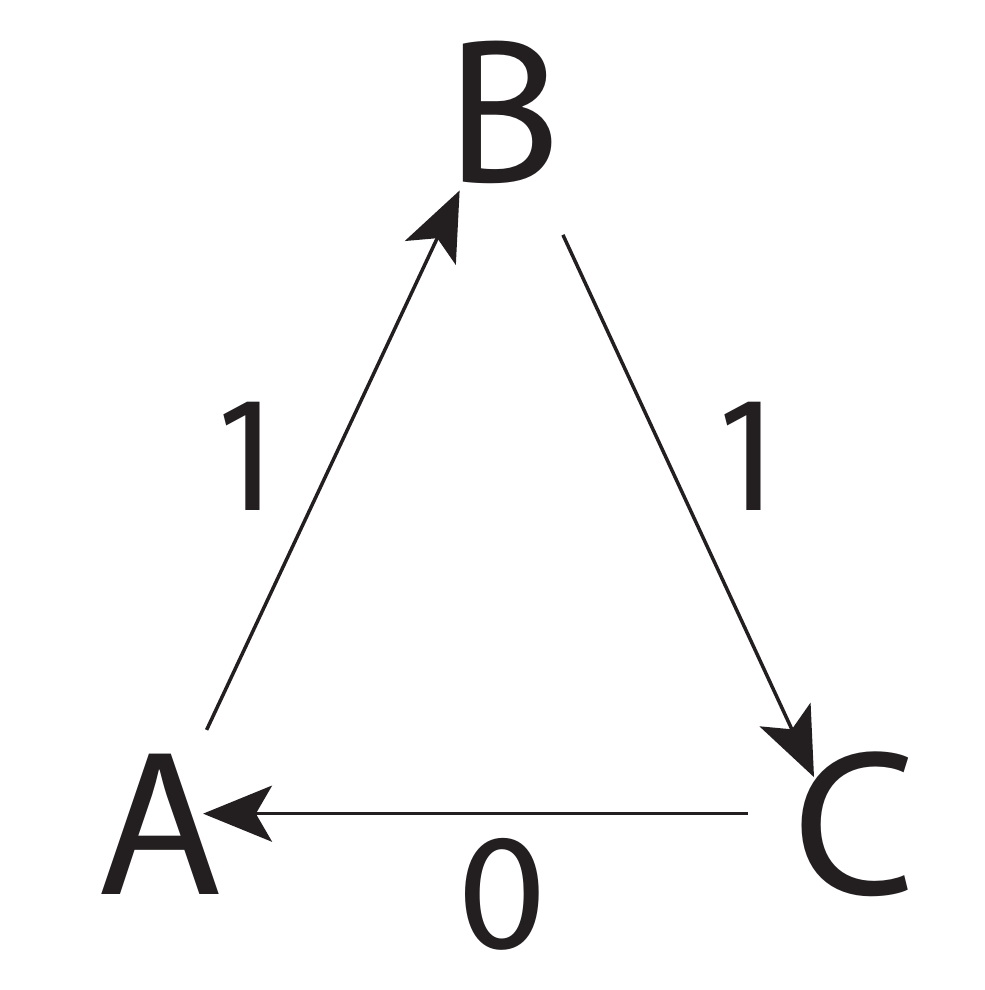}}
\subfigure[C, R, I]{
\label{fig:gr:001} 
\includegraphics[scale=0.19]{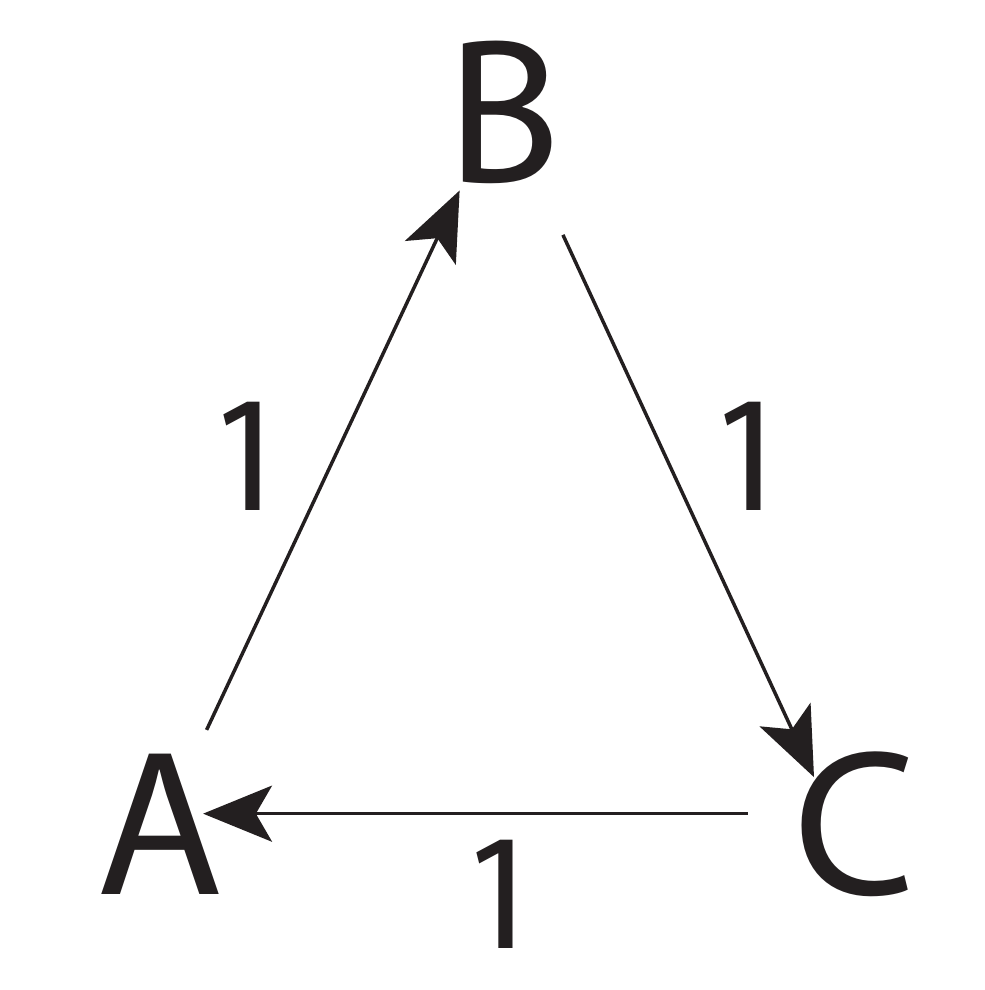}}\\
\subfigure[C, S, T]{
\label{fig:gr:010} 
\includegraphics[scale=0.19]{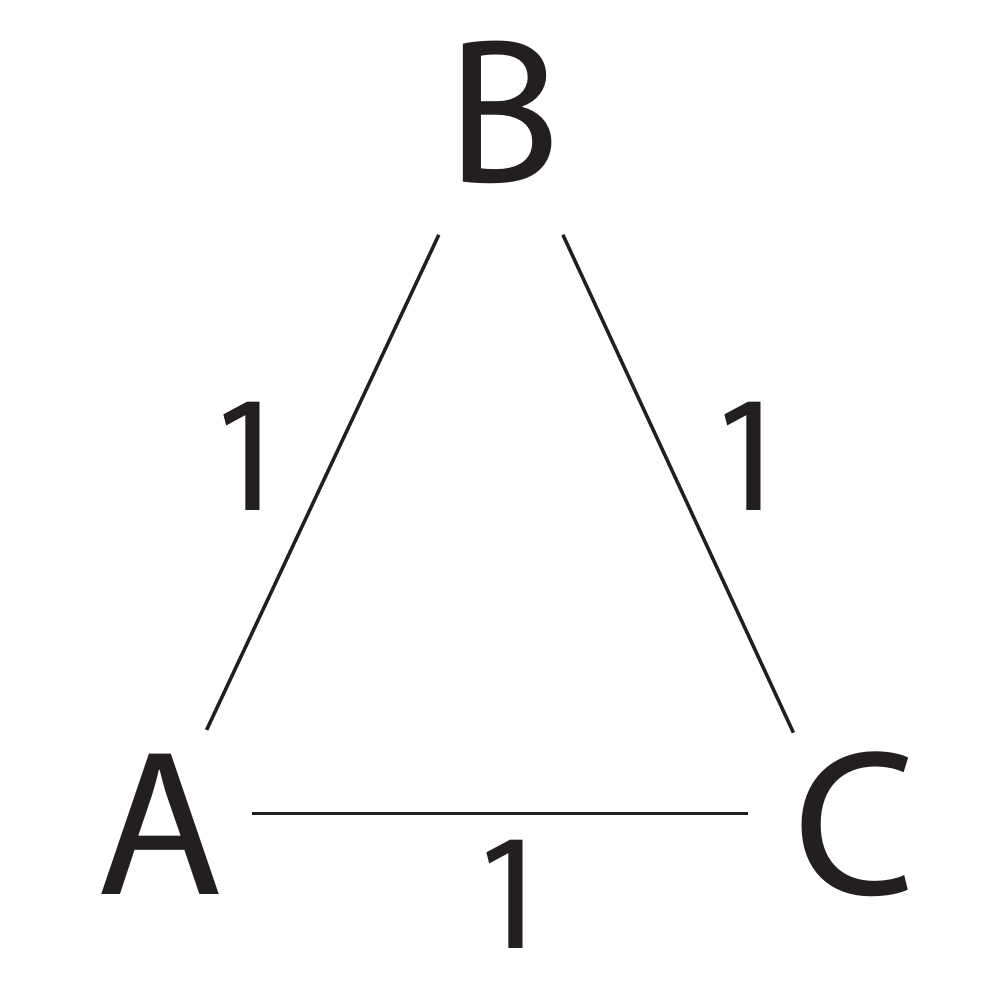}}
\subfigure[C, S, I]{
\label{fig:gr:011} 
\includegraphics[scale=0.19]{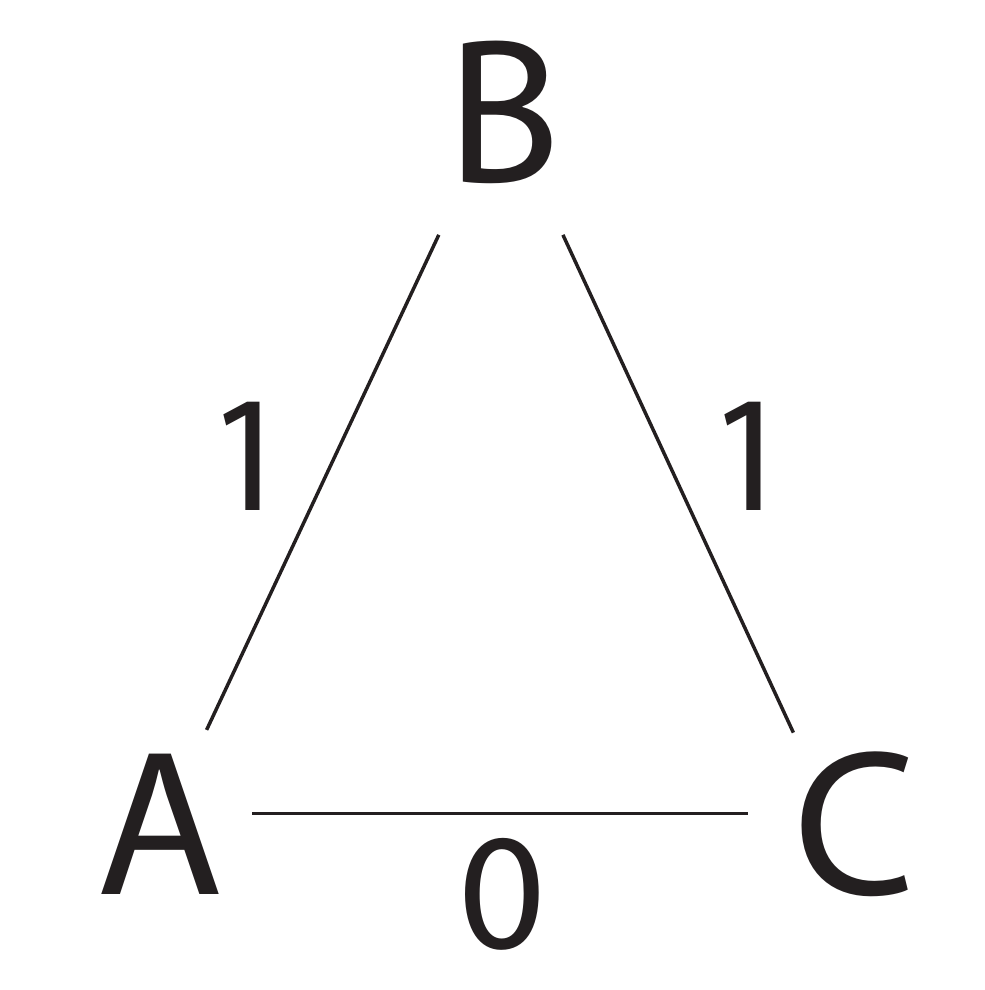}}\\
\subfigure[G, R, T]{
\label{fig:gr:100} 
\includegraphics[scale=0.19]{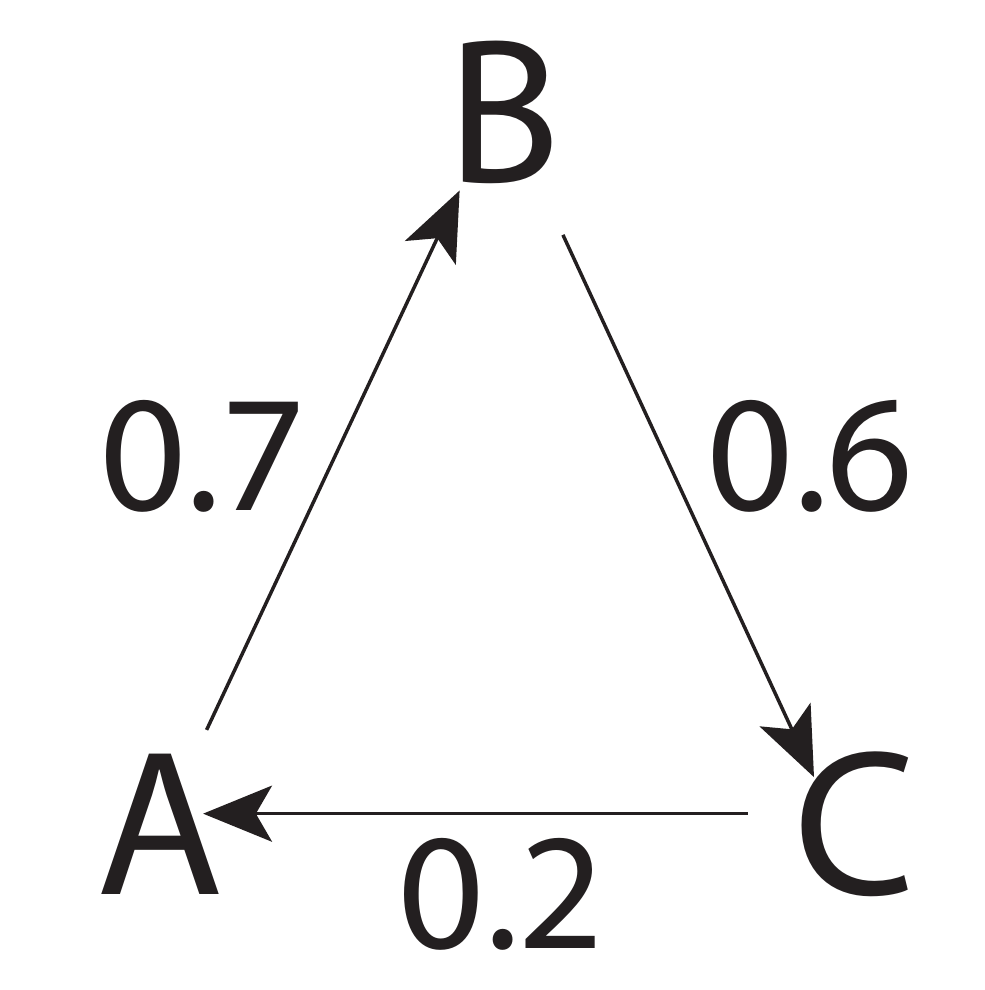}}
\subfigure[G, R, I]{
\label{fig:gr:101} 
\includegraphics[scale=0.19]{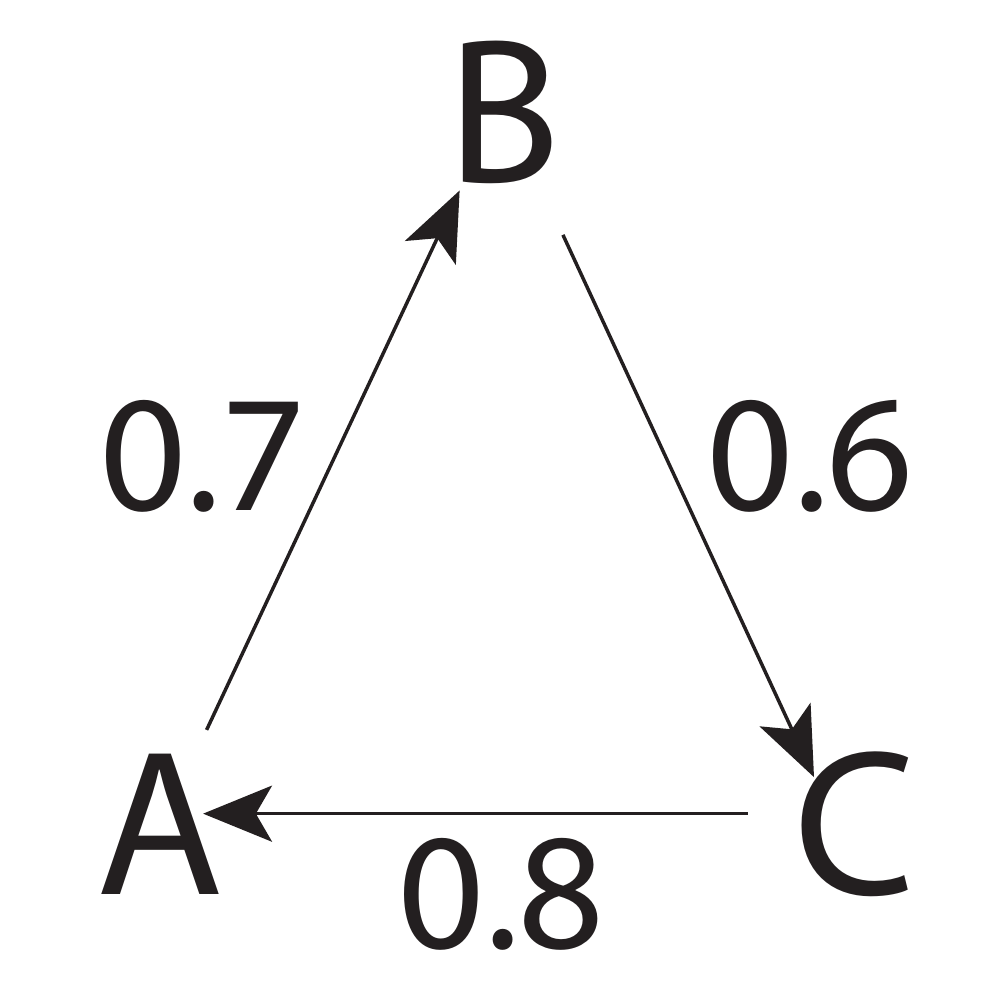}}\\
\subfigure[G, S, T]{
\label{fig:gr:110} 
\includegraphics[scale=0.19]{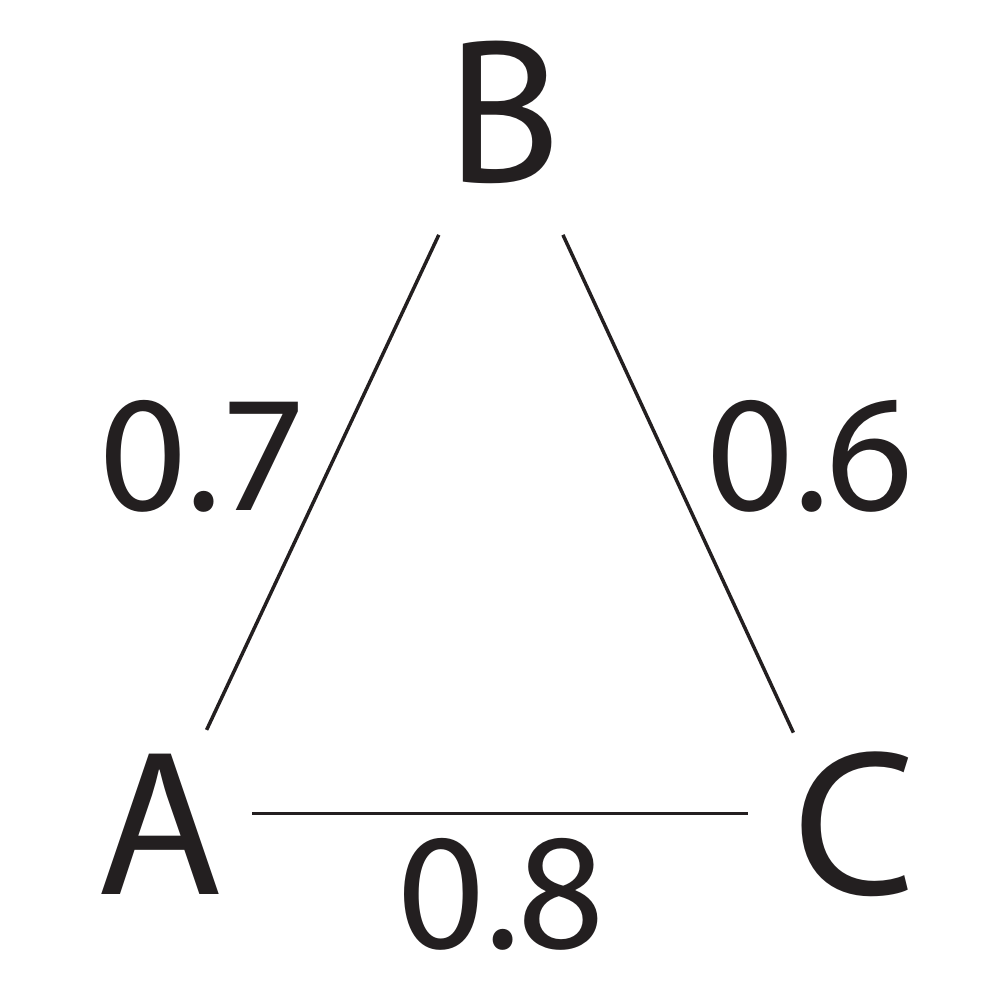}}
\subfigure[G, S, I]{
\label{fig:gr:111} 
\includegraphics[scale=0.19]{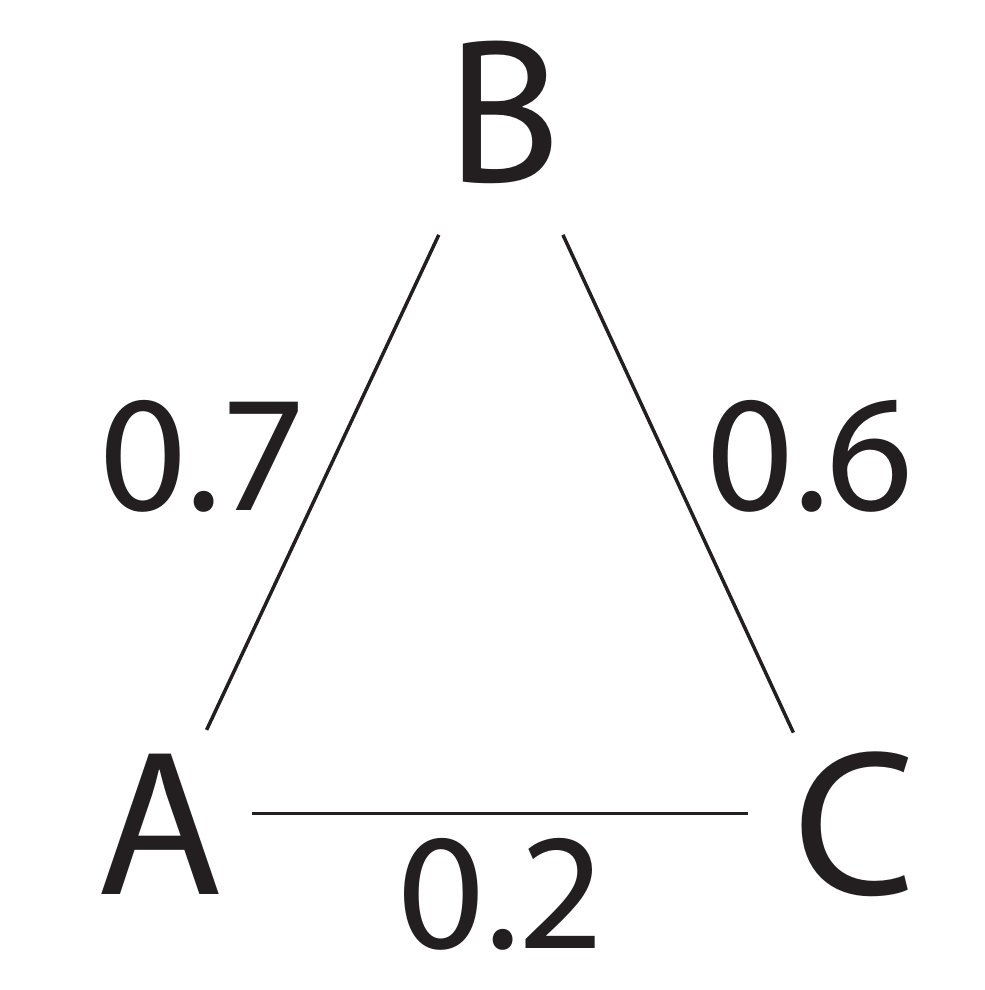}}
\end{minipage}

\caption{Left: example of a multi-graph representing the most general case, where no additional properties of relations are assumed. Right: examples of eight different types of relations in a graph of cardinality three. The following relational properties are illustrated: (C) crisp, (G) graded, (R) reciprocal, (S) symmetric, (T) transitive and (I) intransitive. For the reciprocal relations, (I) refers to a relation that does not satisfy weak stochastic transitivity, while (T) is showing an example of a relation fulfilling strong stochastic transitivity.  For the symmetric relations, (I) refers a relation that does not satisfy $T$-transitivity w.r.t.\ the  \L ukasiewicz t-norm $T_{\bf L}(a,b) = \max(a+b-1,0)$, while (T) is showing an example of a relation that fulfills $T$-transitivity w.r.t.\ the product t-norm $T_{\bf P}(a,b) = ab$. See Section~4 for formal definitions of transitivity.}

\label{fig:examples}
\end{figure}

In this paper we present a general kernel-based approach that unifies all the above cases into one general framework where domain knowledge can be easily specified by choosing a proper kernel and  model structure, while different learning settings are distinguished by means of the loss function. Let $\weightfunc(\node,\node')$ be a binary relation on an object space $\nodeset$, then the following learning settings will be considered in particular:

\begin{itemize*}
\item Crisp relations: when the restriction is made that $Q: \nodeset^2 \rightarrow \{0,1\}$, we arrive at a binary classification task with pairs of objects as input for the classifier.
\item $[0,1]$-valued relations: here it is allowed that relations can take the form $Q: \nodeset^2 \rightarrow [0,1]$, resulting in a regression type of learning setting. The restriction to the interval $[0,1]$ is predominantly made because many mathematical frameworks in fields like fuzzy set theory and decision theory are built upon such relations, using the notion of a fuzzy relation, but in general one can account quite easily for real-graded relations by applying a scaling operation from $\mbr$ to $[0,1]$.
\item Ordinal-valued relations: situated somewhat in the middle between the other two settings, here it is assumed that the actual values of the relation do not matter but rather the provided order information should be learned.
\end{itemize*}
Furthermore, one can integrate different types of domain knowledge in our framework, by guaranteeing that certain properties are satisfied. The following cases can be distinguished:
\begin{itemize*}
\item Symmetric relations. Applications arise in many domains and metric learning or learning similarity measures can be seen as special cases that require additional properties to hold, such as the triangle inequality for metrics and positive definiteness or transitivity properties for similarity measures. As shown below, learning symmetric relations can be interpreted as learning edges in an undirected graph.
\item Reciprocal or antisymmetric relations. Applications arise here in domains such as preference learning, game theory and bioinformatics for representing preference relations, choice probabilities, winning probabilities, gene regulation, etc. We will provide a formal definition below, but, given a rescaling operation from $\mbr$ to $[0,1]$, antisymmetric relations can be converted into reciprocal relations. Similar to symmetric relations, transitivity properties typically guarantee additional constraints that are definitely required for certain applications. It is, for example, well known in decision theory and preference modeling that transitive preference relations result in utility functions \cite{Luce1965,Bodenhofer2007}. Learning reciprocal or antisymmetric relations can be interpreted as learning edges in a directed graph.
\item Ordinary binary relations. Many applications can be found  where neither symmetry nor reciprocity holds. From a graph inference perspective, learning such relations should be seen as learning the edges in a bidirectional graph, where edges in one direction do not impose constraints on edges in the other direction.
\end{itemize*}
Indeed, the framework that we propose below strongly relies on graphs, where nodes represent the data objects that are studied and the edges represent the relations present in the training set. The weights on the edges characterize the values of known relations, while unconnected nodes indicate pairs of objects for which the unknown relation needs to be predicted. The left graph in Figure~\ref{fig:examples} visualizes a toy example representing the most general case where neither symmetry nor reciprocity holds. Depending on the application, the learning algorithm should try to predict the relations for three types of object pairs:
\begin{itemize}
\item pairs of objects that are already present in the training dataset by means of other edges, like the pair (A,B),
\item pairs of objects for which one of the two objects occurs in the training dataset, like the pair (E,F),
\item pairs of objects for which none of the two objects is observed during training, like the pair (F,G).
\end{itemize}
The graphs on the right-hand side in Figure~\ref{fig:examples} show examples of specific types of relations that are covered by our framework. The differences between these relations will become more clear in the following sections.

\section{General framework}
\subsection{Notation and basic concepts}
Let us start with introducing some notations. We assume that the data is structured as a graph $G = (\nodeset,\edgeset,\weightfunc)$, where $\nodeset$ corresponds to the set of nodes $\node$ and $\edgeset \subseteq \nodeset^2$ represents the set of edges $\edge$, for which training labels are provided in terms of relations. Moreover, these relations are represented by training weights $\edgeweight_{\edge}$ on the edges, generated from an unknown underlying relation $\weightfunc: \nodeset^2 \rightarrow [0,1]$. Relations are required to take values in the interval $[0,1]$ because some properties that we need are historically defined for such relations, but an extension to real-graded relations $\hypothesis: \nodeset^2 \rightarrow \mbr$ can always be realized.  Consider $b \in \mbr^+$ and an increasing isomorphism $\sigma : [-b,b] \rightarrow [0,1]$ that satisfies $\sigma(x) = 1 - \sigma(-x)$, then we consider the $\mbr \rightarrow [0,1]$ mapping $\nabla$ defined by:
\begin{eqnarray*}
\nabla(x) & = & \left\{ \begin{array}{ll} 0, &
\textrm{if $ x \le -b$}\\
\sigma(x), &  \textrm{if $-b \le x \le b$} \\
1, & \textrm{if $b \le x$}
\end{array} \right.
\end{eqnarray*}
and its inverse $\nabla^{-1} = \sigma^{-1}$.

Any real-valued relation $\hypothesis: \nodeset^2 \rightarrow \mbr$ can be transformed into a $[0,1]$-valued relation $Q$ as follows:
\begin{eqnarray}
\label{eq:monmap}
Q(\node,\node') = \nabla(h(\node,\node')) \,, \quad \forall (\node,\node') \in \nodeset^2 \,,
\end{eqnarray}
and conversely by means of $\nabla^{-1}$. In what follows we tacitly assume that $\nabla$ has been fixed.

Following the standard notations for kernel methods, we formulate our learning problem as the selection of a suitable function $\hypothesis\in\funspace$, with $\funspace$ a certain hypothesis space, in particular a reproducing kernel Hilbert space (RKHS). More specifically, the RKHS supports in our case hypotheses  $h: \nodeset^2 \rightarrow \mbr$ denoted as
\begin{eqnarray*}
\label{eq:primalmodel}
h(\edge) =  \bm{w}^T \Phi(\edge) \,,
\end{eqnarray*}
with $\bm{w}$ a vector of parameters that needs to be estimated from training data, $\Phi$ a joint feature mapping for edges in the graph (see below) and $\bm{a}^T$ the transpose of a vector $\bm{a}$. Let us denote a training dataset of cardinality $\edgecount = |\edgeset|$ as a set
$T = \{(\edge,\edgeweight_{\edge}) \mid \edge \in \edgeset \}$
of input-label pairs, then we formally consider the following optimization problem, in which we select an appropriate
hypothesis $\hypothesis$ from $\funspace$ for training
data~$T$:
\begin{eqnarray}\label{regalgorithm}
\hat{\hypothesis}=\argmin_{\hypothesis\in\funspace} \frac{1}{q} \sum_{\edge \in \edgeset} \lossfunction(\hypothesis(\edge),\edgeweight_{\edge})
+\regparam\Arrowvert \hypothesis\Arrowvert_{\funspace}^2 \,
\end{eqnarray}
with $\lossfunction$ a given loss function, $\Arrowvert \cdot \Arrowvert_{\funspace}^2$ the traditional quadratic regularizer on the RKHS and $\regparam>0$ a regularization parameter. According to the representer theorem \cite{Scholkopf2002},
any minimizer $\hypothesis\in\funspace$ of
(\ref{regalgorithm}) admits a dual representation of the following form:
\begin{eqnarray}
\label{eq:dualmodel}
\hypothesis(\ol{\edge})
= \hyperplane^T \Phi(\ol{\edge}) = \sum_{\edge \in \edgeset}
\dualpar_{\edge}\kernelf^{\Phi}(\edge,\ol{\edge}) \,,
\end{eqnarray}
with $\dualpar_{\edge} \in\mathbb{R}$ dual parameters, $\kernelf^{\Phi}$ the kernel function associated with the RKHS and $\Phi$ the feature mapping corresponding to $\kernelf^{\Phi}$ and
\begin{eqnarray*}
\hyperplane=\sum_{\edge \in \edgeset}
\dualpar_{\edge}\Phi(\edge).
\end{eqnarray*}
We will alternate several times between the primal and dual representation for $\hypothesis$ in the remainder of this article.

The primal representation as defined in (\ref{eq:primalmodel}) and its dual equivalent (\ref{eq:dualmodel}) yield an RKHS defined on edges in the graph. In addition, we will establish an RKHS defined on nodes, as every edge consists of a couple of nodes.  Given an input space $\mathcal{V}$ and a kernel $\kernelf:\mathcal{V}\times\mathcal{V}\rightarrow\mathbb{R}$, the RKHS associated with $\kernelf$ can be considered as the completion of
\begin{equation*}
\left\{ f \in \mathbb{R}^\mathcal{V}  \left\arrowvert
f(\node)=\sum_{i=1}^m\beta_i\kernelf(\node,\node_i)\right.\right\},
\end{equation*}
in the norm
\[
\Arrowvert f\Arrowvert_\kernelf
=\sqrt{\sum_{i,j}\beta_i\beta_j\kernelf(\node_i,\node_j)},
\]
where $\beta_i\in\mathbb{R},m\in\mathbb{N},\node_i\in\mathcal{V}$.

\subsection{Learning arbitrary relations}
As mentioned in the introduction, both crisp and graded relations can be handled by our framework. To make a subdivision between different cases, a loss function needs to be specified. For crisp relations, one can typically use the hinge loss, which is given by:
\begin{eqnarray*}
\lossfunction(\hypothesis(\edge),\edgeweight) = [1- \edgeweight \hypothesis(\edge)]_+ \,,
\end{eqnarray*}
with $[\cdot]_+$ the positive part of the argument. Alternatively, one can opt to optimize a probabilistic loss function like the logistic loss:
\begin{eqnarray*}
\lossfunction(\hypothesis(\edge),\edgeweight) = \ln(1+\exp(-\edgeweight \hypothesis(\edge))) \,.
\end{eqnarray*}
Conversely, if in a given application the observed relations are graded instead of crisp, other loss functions have to be considered. Hence, we will run experiments with a least-squares loss function:
\begin{equation}\label{regrloss}
\lossfunction(\hypothesis(\edge),\edgeweight) = (\edgeweight_\edge-\hypothesis(\edge))^2 \,,
\end{equation}
resulting in a regression type of learning setting. Alternatively, one could prefer to optimize a more robust regression loss like the $\epsilon$-insensitive loss, in case outliers are expected in the training dataset.

So far, our framework does not differ from standard classification and regression algorithms. However, the specification of a more precise model structure for (\ref{eq:primalmodel}) offers a couple of new challenges. In the most general case, when no further restrictions on the underlying relation can be specified, the following Kronecker product feature mapping is proposed to express pairwise interactions between features of nodes:
\begin{eqnarray*}
\Phi(\edge) = \Phi(\node,\node') = \phi(\node) \otimes \phi(\node')\,,
\end{eqnarray*}
where $\phi$ represents the feature mapping for individual nodes. A formal definition of the Kronecker product can be found in the appendix.
As first shown in \cite{Ben-Hur2005}, the Kronecker product pairwise feature mapping yields the Kronecker product edge kernel (a.k.a.\ the tensor product pairwise kernel) in the dual representation:
\begin{eqnarray}
\label{eq:tppk}
\kernelf_{\otimes}^{\Phi}(\edge,\ol{\edge}) = \kernelf_{\otimes}^{\Phi}(\node,\node',\ol{\node},\ol{\node}')= \kernelf^{\phi}(\node,\ol{\node}) \kernelf^{\phi}(\node',\ol{\node}') \,,
\end{eqnarray}
with $K^{\phi}$ the kernel corresponding to $\phi$.

This section aims to formally prove that the Kronecker product edge kernel is the best kernel one can choose, when no further domain knowledge is provided about the underlying relation that generates the data. We claim that with an appropriate choice for $K^{\phi}$, such as the Gaussian RBF kernel, the kernel $K^{\Phi}$ generates a class $\funspace$ of universally approximating functions for learning any type of relation.
Armed with the definition of universality for kernels and the Stone-Weierstra{\ss} theorem \cite{Steinwart2002consistency}, we arrive at the following theorem concerning the Kronecker product pairwise kernels:
\begin{theorem}\label{unikrontheorem}
Let us assume that the space of nodes $\nodeset$ is a compact metric space.
If a continuous kernel $\kernelf^\phi$ is universal on $\nodeset$, then $\kernelf_{\otimes}^\Phi$ defines a universal kernel on $\edgeset$.
\end{theorem}

The proof can be found in the appendix. We would like to emphasize that one cannot conclude from the theorem that the Kronecker product pairwise kernel is the best kernel to use in all possible situations. The theorem only shows that the Kronecker product pairwise kernel makes a reasonably good choice, if no further domain knowledge about the underlying relation is known.
Namely, the theorem says that given a suitable sample of data, the RKHS of the kernel contains functions that are arbitrarily close to any continuous relation in the uniform norm. However, the theorem does not say anything about how likely it is to have, as a training set, such a data sample that can represent the approximating function. Further, the theorem only concerns graded relations that are continuous and therefore crisp relations and graded, discontinuous relations require more detailed considerations.

Other kernel functions might of course outperform the Kronecker product pairwise kernel in applications where domain knowledge can be incorporated in the kernel function. In the following section we discuss reciprocity, symmetry and transitivity as three relational properties that can be represented by means of more specific kernel functions. As a side note, we also introduce the Cartesian pairwise kernel, which is formally defined as follows
$$K_{C}^{\Phi}(\node,\node',\ol{\node},\ol{\node}') = K^{\phi}(\node', \ol{\node}') [\node = \ol{\node}] + K^{\phi}(\node,\ol{\node}) [\node' = \ol{\node}'] \,,$$
with $[.]$ the indicator function, returning one when both elements are identical and zero otherwise. This kernel was recently proposed by \cite{Kashima2009} as an alternative to the Kronecker product pairwise kernel. By construction, the Cartesian pairwise kernel has important limitations, since it cannot generalize to couples of nodes for which both nodes did not appear in the training dataset.

\section{Special relations}
Thus, if no further information is available about the relation that underlies the data, one should definitely use the Kronecker product edge kernel.
In this most general case, we allow that for any pair of nodes in the graph several edges can exist, in which an edge in one direction does not necessarily impose constraints on the edge in the opposite direction. Multiple edges in the same direction can connect two nodes, leading to a multi-graph as in Figure~\ref{fig:examples}, where two different edges in the same direction connect nodes $D$ and $E$. This construction is required to allow repeated measurements. However, two important subclasses of relations deserve further attention: reciprocal relations and symmetric relations.

\subsection{Reciprocal relations}
This subsection briefly summarizes our previous work on learning reciprocal relations \cite{Pahikkala2010}. Let us start with a definition of this type of relation.
\begin{definition}
A binary relation $\weightfunc: \nodeset^2 \rightarrow [0,1]$ is called a reciprocal relation if for all $(\node,\node') \in \nodeset^2$ it holds that $\weightfunc(\node,\node') = 1 - \weightfunc(\node',\node)$. \end{definition}
\begin{definition}
A binary relation $\hypothesis: \nodeset^2 \rightarrow \mbr$ is called an antisymmetric relation if for all $(\node,\node') \in \nodeset^2$ it holds that $\hypothesis(\node,\node') = - \hypothesis(\node',\node)$. \end{definition}
For reciprocal and antisymmetric relations, every edge $\edge=(\node,\node')$ in a multi-graph like Figure~\ref{fig:examples} induces an unobserved invisible edge $\edge_R = (\node',\node)$ with appropriate weight in the opposite direction. The transformation operator $\nabla$ transforms an antisymmetric relation into a reciprocal relation. Applications of reciprocal relations arise here in domains such as preference learning, game theory and bioinformatics for representing preference relations, choice probabilities, winning probabilities, gene regulation, etc. The weight on the edge defines the real direction of such an edge. If the weight on the edge $\edge = (\node,\node')$ is higher than 0.5, then the direction is from $v$ to $v'$, but when the weight is lower than 0.5, then the direction should be interpreted as inverted, for example, the edges from $A$ to $C$ in Figures~\ref{fig:examples} (a) and (e) should be interpreted as edges starting from $A$ instead of $C$. If the relation is $3$-valued as $Q: \nodeset^2 \rightarrow \{0,1/2,1\}$, then we end up with a three-class ordinal regression setting instead of an ordinary regression setting.

Interestingly, reciprocity can be easily incorporated in our framework.

\begin{proposition}
Let $\Psi$ be a feature mapping on $\nodeset^2$ and let $\hypothesis$ be a hypothesis defined by (\ref{eq:primalmodel}), then the relation $Q$ of type (\ref{eq:monmap}) is reciprocal if $\Phi$ is given by $$\Phi_R(\edge)=\Phi_R(\node,\node')=\Psi(\node,\node')-\Psi(\node',\node) \,.$$
\end{proposition}

The proof is immediate. In addition, one can easily show that reciprocity as domain knowledge can be enforced in the dual formulation. Let us in the least restrictive form now consider the Kronecker product for $\Psi$, then one obtains for $\Phi_R$ the kernel $K_{\otimes R}^{\Phi}$ given by $K_{\otimes R}^{\Phi}(\edge,\ol{\edge})=$
\begin{eqnarray}
\label{eq:recedgekernel}
2 \big(K^{\phi}(\node,\ol{\node}) K^{\phi}(\node',\ol{\node}') - K^{\phi}(\node,\ol{\node}') K^{\phi}(\node',\ol{\node})\big)\,.
\end{eqnarray}
The following theorem shows that this kernel can represent any type of reciprocal relation.
\begin{theorem}\label{antisymmetrictheorem}
Let
\begin{equation*}
R(\nodeset^2)=\{ t \mid t \in C(\nodeset^2),t(\node,\node')= - t(\node',\node)\}
\end{equation*}
be the space of all continuous antisymmetric relations from $\nodeset^2$ to $\mathbb{R}$.
If $K^\phi$ on $\nodeset$ is universal, then for every function $t\in R(\nodeset^2)$ and every $\epsilon > 0$, there exists a function $h$ in the RKHS induced by the kernel $K_{\otimes R}^{\Phi}$ defined in (\ref{eq:recedgekernel}), such that
\begin{equation}\label{recclaim}
\max_{(\node,\node')\in \nodeset^2}\left\{\left\arrowvert t(\node,\node')-\hypothesis(\node,\node')\right\arrowvert\right\}\leq\epsilon \,.
\end{equation}
\end{theorem}
The proof can be found in the appendix.

\subsection{Symmetric relations}
Symmetric relations form another important subclass of relations in our framework. As a specific type of symmetric relations, similarity relations constitute the underlying relation in many application domains where relations between objects need to be learned. Symmetric relations are formally defined as follows.
\begin{definition}
A binary relation $\weightfunc: \nodeset^2 \rightarrow [0,1]$ is called a symmetric relation if for all $(\node,\node') \in \nodeset^2$ it holds that $\weightfunc(\node,\node') = \weightfunc(\node',\node)$.
\end{definition}
\begin{definition}
A binary relation $\hypothesis: \nodeset^2 \rightarrow \mbr$ is called a symmetric relation if for all $(\node,\node') \in \nodeset^2$ it holds that $\hypothesis(\node,\node') = \hypothesis(\node',\node)$.
\end{definition}
Note that $\nabla$ preserves symmetry. For symmetric relations, edges in multi-graphs like Figure~\ref{fig:examples} become undirected. Applications arise in many domains and metric learning or learning similarity measures can be seen as special cases. If the relation is $2$-valued as $Q: \nodeset^2 \rightarrow \{0,1\}$, then we end up with a classification setting instead of a regression setting.

Just like reciprocal relations, it turns out that symmetry can be easily incorporated in our framework.
\begin{proposition}
Let $\Psi$ be a feature mapping on $\nodeset^2$ and let $\hypothesis$ be a hypothesis defined by (\ref{eq:primalmodel}), then the relation $Q$ of type (\ref{eq:monmap}) is symmetric if $\Phi$ is given by $$
\Phi_S(\edge)=\Phi_S(\node,\node')=\Psi(\node,\node')+\Psi(\node',\node) \,.$$
\end{proposition}
In addition, by using mathematical properties of the Kronecker product, one obtains in the dual formulation an edge kernel that looks very similar to the one derived for reciprocal relations.  Let us again consider the Kronecker product for $\Psi$, then one obtains for $\Phi_S$ the kernel $K_{\otimes S}^{\Phi}$ given by $K_{\otimes S}^{\Phi}(\edge,\ol{\edge})=$
\begin{eqnarray*}
2 \big(K^{\phi}(\node,\ol{\node}) K^{\phi}(\node',\ol{\node}') + K^{\phi}(\node,\ol{\node}') K^{\phi}(\node',\ol{\node})\big) \,.
\end{eqnarray*}
Thus, the substraction of kernels in the reciprocal case becomes an addition of kernels in the symmetric case. The above kernel has been used for predicting protein-protein interactions in bioinformatics \cite{Ben-Hur2005}  and it has been theoretically analyzed in \cite{Hue2010}. More specifically, for some methods one has shown in the latter paper that enforcing symmetry in the kernel function yields identical results as adding every edge twice to the dataset, by taking each of the two nodes once as first element of the edge.  Unlike many existing kernel-based methods for pairwise data, the models obtained with these kernels are able to represent any reciprocal or symmetric relation respectively, without imposing additional transitivity properties of the relations.

We also remark that for symmetry as well, one can prove that the Kronecker product edge kernel yields a model that is flexible enough to represent any type of underlying relation.
\begin{theorem}
Let
\begin{equation*}
S(\nodeset^2)=\{t \mid t\in C(\nodeset^2),t(\node,\node')=t(\node',\node)\}
\end{equation*}
be the space of all continuous symmetric relations from $\nodeset^2$ to $\mathbb{R}$.
If $K^\phi$ on $\nodeset$ is universal, then for every function $t\in S(\nodeset^2)$ and every $\epsilon > 0$, there exists a function $h$ in the RKHS (\ref{eq:primalmodel}) induced by the kernel (\ref{eq:recedgekernel}), such that
\begin{equation*}
\max_{(\node,\node')\in \nodeset^2}\left\{\left\arrowvert t(\node,\node')-\hypothesis(\node,\node')\right\arrowvert\right\}\leq\epsilon.
\end{equation*}
\end{theorem}
The proof is analogous to that of Theorem~\ref{antisymmetrictheorem} (see appendix).

As a side note, we remark that a symmetric and reciprocal version of the Cartesian kernel can be introduced as well.

\section{Relationships with fuzzy set theory}
The previous section revealed that specific Kronecker product edge kernels can be constructed for modeling reciprocal and symmetric relations, without requiring any further background about these relations. In this section we demonstrate that the Kronecker product edge kernels $K_{\otimes}^{\Phi}$, $K_{\otimes R}^{\Phi}$ and $K_{\otimes S}^{\Phi}$ are particularly useful for modeling intransitive relations. Intransitive relations occur in a lot of real-world scenarios, like game playing \cite{DeSchuymer2003,Fisher2008}, competition between bacteria \cite{Kerr2002,czaran2002chemical,nowak2002,Kirkup2004,karolyi2005rps,reichenbach2007} and fungi \cite{Boddy2000}, mating choice of lizards \cite{Sinervo1996} and food choice of birds \cite{Waite2001}, to name just a few. In an informal way, Figure~\ref{fig:examples} shows with the help of examples what transitivity means for symmetric and reciprocal relations that are crisp and graded.

Despite the occurrence of intransitive relations in many domains, one has to admit that most applications are still characterized by relations that fulfill relatively strong transitivity requirements. For example, in decision making, preference modeling and social choice theory, one can argue that reciprocal relations like choice probabilities and preference judgments should satisfy certain transitivity properties, if they represent rational human decisions made after well-reasoned comparisons on objects \cite{Luce1965,Fishburn1991,Tversky1998}. For symmetric relations as well, transitivity plays an important role \cite{Gower1986,Jakel2008}, when modeling similarity relations, metrics, kernels, etc.

It is for this reason that transitivity properties have been studied extensively in fuzzy set theory and related fields. For reciprocal relations, one traditionally uses the notion of stochastic transitivity \cite{Luce1965}.
\begin{definition}
Let $g$ be an increasing $[1/2,1]^2 \rightarrow [0,1]$ mapping. A
reciprocal relation $Q: \nodeset^2 \rightarrow [0,1]$ is called
$g$-stochastic transitive if for any $(\node_1,\node_2,\node_3)
\in \nodeset^3$
\begin{eqnarray*}
\big( Q(\node_1,\node_2) \geq 1/2 \wedge Q(\node_2,\node_3) \geq 1/2 \big)
\Rightarrow Q(\node_1,\node_3) \geq
g(Q(\node_1,\node_2),Q(\node_2,\node_3))\,.
\end{eqnarray*}
\end{definition}
Important special cases are weak stochastic transitivity when $g(a,b) =
1/2$, moderate stochastic transitivity when $g(a,b) =
\min(a,b)$ and strong stochastic transitivity when $g(a,b) =
\max(a,b)$.
Alternative (and more general) frameworks are FG-transitivity \cite{Switalski2003} and cycle transitivity \cite{DeBaets2005,DeBaets2006}. For graded symmetric relations, the notion of $T$-transitivity has been put forward \cite{DeBaets2002,Moser2006}.
\begin{definition}
A symmetric relation $Q: \nodeset^2 \rightarrow [0,1]$ is called
$T$-transitive with $T$ a t-norm if for any
$(\node_1,\node_2,\node_3) \in \nodeset^3$
\begin{eqnarray}
T(Q(\node_1,\node_2),Q(\node_2,\node_3)) \leq
Q(\node_1,\node_3)\,.
\end{eqnarray}
\end{definition}
Three important t-norms
are the minimum t-norm $T_{\bf M}(a,b) = \min(a,b)$, the product
t-norm $T_{\bf P}(a,b) = ab$ and the \L ukasiewicz t-norm $T_{\bf
L}(a,b) = \max(a+b-1,0)$.

In addition, several authors have shown that various forms of transitivity give rise to utility representable or numerically representable relations, also called fuzzy weak orders -- see e.g.\ \cite{Luce1965,Billot1995,Koppen1995,Fono2007,Bodenhofer2007}. We will use the term ranking representability to establish a link with machine learning. We give a slightly specific definition that unifies reciprocal and symmetric relations.
\begin{definition}
A reciprocal or symmetric relation $\weightfunc: \nodeset^2 \rightarrow [0,1]$ is called ranking representable if there exists a ranking function $\rankfunc: \nodeset \rightarrow \mbr$ such that for all $(\node,\node')
\in \nodeset^2$ it respectively holds that
\begin{enumerate*}
\item $\weightfunc(\node,\node') = \nabla (\rankfunc(\node) - \rankfunc(\node')) \,$ (reciprocal case) ;
\item $\weightfunc(\node,\node') = \nabla (\rankfunc(\node) + \rankfunc(\node')) \,$ (symmetric case) .
\end{enumerate*}
\end{definition}
The main idea is that ranking representable relations can be constructed from a utility function $f$. Ranking representable reciprocal relations correspond to directed acyclic graphs, and a unique ranking of the nodes in such graphs can be obtained with topological sorting algorithms. The ranking representable reciprocal relations of  Figures~\ref{fig:examples} (a) and (e) for example yield the global ranking $A \succ B \succ C$. Interestingly, ranking representability of reciprocal relations and symmetric relations can be easily achieved in our framework by simplifying the joint feature mapping $\Psi$. Let $\Psi(\node,\node') = \phi(\node)$ such that $K^{\Phi}$ simplifies to
\begin{eqnarray*}
\arraycolsep=2pt
K_{\rankfunc R}^{\Phi}(\edge,\ol{\edge})&=&K^{\phi}(\node,\ol{\node}) + K^{\phi}(\node',\ol{\node}')  -K^{\phi}(\node,\ol{\node}') -
K^{\phi}(\node',\ol{\node}) \,, \\
K_{\rankfunc S}^{\Phi}(\edge,\ol{\edge})&=&K^{\phi}(\node,\ol{\node}) + K^{\phi}(\node',\ol{\node}') +K^{\phi}(\node,\ol{\node}') +
K^{\phi}(\node',\ol{\node}) \,,
\end{eqnarray*}
when $\Phi(\node,\node') = \Phi_R(\node,\node')$ or $\Phi(\node,\node') = \Phi_S(\node,\node')$, respectively, then the following proposition holds.
\begin{proposition}
The relation $Q: \nodeset^2 \rightarrow [0,1]$ given by (\ref{eq:monmap}) and $h$ defined by (\ref{eq:primalmodel}) with $K^{\Phi} = K_{\rankfunc R}^{\Phi}$ (respectively $K^{\Phi} = K_{\rankfunc S}^{\Phi}$) is a ranking representable reciprocal (respectively symmetric) relation.
\end{proposition}
The proof directly follows from the fact that for this specific kernel, $\hypothesis(\node,\node')$ can be respectively written as $\rankfunc(\node) - \rankfunc(\node')$ and $\rankfunc(\node) + \rankfunc(\node')$. The kernel $K_{\rankfunc R}^{\Phi}$ has been initially introduced in \cite{Herbrich2000} for ordinal regression and during the last decade it has been extensively used as a main building block in many kernel-based ranking algorithms. Since ranking representability of reciprocal relations implies strong stochastic transitivity of reciprocal relations, $K_{\rankfunc R}^{\Phi}$ can represent this type of domain knowledge.

The notion of ranking representability is powerful for reciprocal relations, because the majority of reciprocal relations satisfy this property, but for symmetric relations it has a rather limited applicability. Ranking representability as defined above cannot represent relations that originate from an underlying metric or similarity measure. For such relations, one needs another connection with its roots in Euclidean metric spaces \cite{Gower1986}.
\begin{definition}
A symmetric relation $\weightfunc: \nodeset^2 \rightarrow [0,1]$ is called Euclidean representable if there exists a ranking function $f: \nodeset \rightarrow \mbr$ such that for all pairs $(\node,\node')
\in \nodeset^2$ it holds that
\begin{eqnarray}
\label{eq:rrsym}
\weightfunc(\node,\node') = \nabla ((\rankfunc(\node) - \rankfunc(\node'))^T(\rankfunc(\node) - \rankfunc(\node'))) \,,
\end{eqnarray}
with $\vec{a}^T$ the transpose of a vector $\vec{a}$.
\end{definition}

Euclidean representability as defined here basically can be seen as Euclidean embedding or Multidimensional Scaling in a $z$-dimensional space \cite{Zhang2003}. In its most restrictive form, when $z=1$, it implies that the symmetric relation can be constructed from the Euclidean distance in a one-dimensional space. When such a one-dimensional embedding can be realized, one global ranking of the objects can be found, similar to reciprocal relations.  Nevertheless, although models of type (\ref{eq:rrsym}) with $z=1$ are sometimes used in graph inference \cite{Vert2005} and semi-supervised learning \cite{Belkin2006}, we believe that situations where symmetric relations become Euclidean representable in a one-dimensional space occur very rarely, in contrast to reciprocal relations. The extension to $z > 1$ on the other hand does not guarantee the existence of one global ranking, then Euclidean representability still enforces some interesting properties, because it guarantees that the relation $Q$ is constructed from a Euclidean metric space with a dimension upper bounded by the number of nodes $\nodecount$. Moreover, this type of domain knowledge about relations can be incorporated in our framework. To this end, let $\Phi(\node,\node') = \Phi_S(\node,\node')$ and let $\Psi(\node,\node') = \phi(\node) \otimes (\phi(\node) - \phi(\node'))$ such that $K^{\Phi}$ becomes
\begin{eqnarray*}
\arraycolsep=2pt
K_{\rm MLPK}^{\Phi}(\edge,\ol{\edge}) &=& (K_{fR}^{\Phi}(\edge,\ol{\edge}))^2 \\
&=& \big(K^{\phi}(\node,\ol{\node}) + K^{\phi}(\node',\ol{\node}') -K^{\phi}(\node,\ol{\node}') -
K^{\phi}(\node',\ol{\node})\big)^2 \,.
\end{eqnarray*}
This kernel has been called the metric learning pairwise kernel by \cite{Vert2007}.
As a consequence, the vector of parameters $\hyperplane$ can be rewritten as an $\feaspacecount \times \feaspacecount$ matrix $\bm{W}$ where $\bm{W}_{ij}$ corresponds to the parameter associated with $(\phi_i(\node) - \phi_i(\node'))(\phi_j(\node) - \phi_j(\node'))$ such that $\bm{W}_{ij} = \bm{W}_{ji}$.
\begin{proposition}\label{thm:mlpk}
If $\bm{W}$ is positive semi-definite, then the symmetric relation $Q: \nodeset^2 \rightarrow [0,1]$ given by (\ref{eq:monmap}) with $h$ defined by (\ref{eq:primalmodel}) and $K^{\Phi} = K_{\rm MLPK}^{\Phi}$ is an Euclidean representable symmetric relation.
\end{proposition}
See the appendix for the proof. Although the model established by $K_{\rm MLPK}^{\Phi}$ does not result in a global ranking, this model strongly differs from the one established with $K_{\otimes S}^{\Phi}$, since $K_{\rm MLPK}^{\Phi}$ can only represent symmetric relations that exhibit transitivity properties. Therefore, one should definitely use $K_{\rm MLPK}^{\Phi}$ when, for example, the underlying relation corresponds to a metric or a similarity relation, while the kernel $K_{\otimes S}^{\Phi}$ should be preferably used for symmetric relations for which no further domain knowledge can be assumed beforehand.

\section{Relationships with other machine learning algorithms}
As explained in Section~2, the transition from a standard classification or regression setting to the setting of learning graded relations should be rather found in the specification of joint feature mappings over couples of objects, thereby naturally leading to the introduction of specific kernels. Any existing machine learning algorithm for classification or regression can in principle be adopted if joint feature mappings are constructed explicitly. Since kernel methods avoid this explicit construction, they can often outperform non-kernelized algorithms in terms of computational efficiency \cite{Scholkopf2002}. As a second main advantage, kernel methods allow to express similarity scores for structured objects, such as strings, graphs and trees and text \cite{Shawetaylor2004}. In our setting of learning graded relations, this implies that one should plug these domain-specific kernel functions into (\ref{eq:tppk}) or the other pairwise kernels that are discussed in this paper. Such a scenario is in fact common practice in some applications of Kronecker product pairwise kernels, such as predicting protein-ligand compatibility in bioinformatics \cite{Jacob2008}. String kernels or graph kernels can be defined on various types of biological structures \cite{Vishwanathan2010} and Kronecker product pairwise kernels then combine these object-based kernels into relation-based kernels (thus, node kernels versus edge kernels).

The edge kernels we discussed in this article can be utilized within a wide variety of kernel methods. Since we focus on learning graded relations, one naturally arrives at a regression setting. In the following section, we run some experiments with regularized least-squares methods, which optimize (\ref{regrloss}) using a hypothesis space induced by kernels. The solution is found by simply solving a system of linear equations \cite{Saunders1998,Suykens2002,Shawetaylor2004,pahikkala2009preferences}.

Apart from kernel methods, we briefly mention a number of other algorithms that are somewhat connected, even though they provide solutions for different learning problems. If pairwise relations are considered between objects of two different domains, one arrives at a learning setting that is referred to as predicting labels for dyadic data \cite{Menon2010}. Examples of such settings include link prediction in bipartite graphs and movie recommendation for users. As such, one could also argue that specific link prediction and matrix factorization methods could be applied in our setting as well, see e.g.\ \cite{Srebro2005,Miller2009,Lawrence2009}. However, these methods have been primarily designed for exploiting relationships in the output space, whereas feature representations of the objects are often not observed or simply irrelevant. Moreover, similar to the Cartesian pairwise kernel, these methods cannot be applied in situations where predictions need to be made for two new nodes that were not present in the training dataset.

Another connection can be observed with multivariate regression and structured output prediction methods. Such methods have been occasionally applied in settings where relations had to be learned \cite{Geurts2007}. Also recall that structured output prediction methods use Kronecker product pairwise kernels on a regular basis to define joint feature representations of inputs and outputs \cite{Tsochantaridis2005, Weston2007}.

In addition to predictive models for dyadic data, one can also detect connections with certain information retrieval and pattern matching methods. However, these methods predominantly use similarity as underlying relation,
often in a purely intuitive manner, as a nearest neighbor type of learning, so they can be considered as much more restrictive.
Consider the example of protein ranking
\cite{Weston2004} or algorithms like \emph{query by document} \cite{Yang2009}. These methods simply look for rankings where the most similar objects w.r.t.\ the query object appear on top, contrary to our approach, which should be considered as much more general, since we learn rankings from any type of binary relation. Nonetheless, similarity relations will of course still occupy a prominent place in our framework as an important special case.

\section{Experiments}
In the experiments, we test the ability of the pairwise kernels to model different types of relations, and the effect of enforcing prior knowledge about the properties of the learned relations. To this end, we train the regularized least-squares (RLS) algorithm to regress the relation values \cite{pahikkala2009preferences}.  We perform experiments on both symmetric and reciprocal relations, considering both synthetic and real-world data. In addition to the standard, symmetric and reciprocal Kronecker product pairwise kernels, we also consider the Cartesian kernel, the symmetric Cartesian kernel and the metric learning pairwise kernel.

\begin{table}
\centering
\begin{tabular}{ll}
\hline
Abbreviation & Method\\
\hline
\hline
MPRED & Predicting the mean \\
$K_{\otimes}^{\Phi}$ & Kronecker Product Pairwise Kernel\\
$K_{\otimes S}^{\Phi}$ & Symmetric Kronecker Product Pairwise Kernel\\
$K_{\otimes R}^{\Phi}$ & Reciprocal Kronecker Product Pairwise Kernel\\
$K_{\rm MLPK}^{\Phi}$ & Metric Learning Pairwise Kernel\\
$K_{C}^{\Phi}$ & Cartesian Product Pairwise Kernel\\
$K_{C S}^{\Phi}$ & Symmetric Cartesian Pairwise Kernel\\
\hline
\end{tabular}
\centering
\caption{Methods considered in the experiments}
\label{abbreviations}
\end{table}


\subsection{Synthetic data: learning similarity measures}
Experiments on synthetic data were conducted to illustrate the behavior of the different kernels in terms of the transitivity of the relation to be learned. A parametric family of cardinality-based similarity measures for sets was considered as the relation of interest \cite{DeBaets2001}. For two sets $A$ and $B$, let us define the following cardinalities:
\begin{eqnarray*}
\arraycolsep=2pt
\Delta_{A,B} &=& |A \setminus B| + |B \setminus A| \,, \\
\delta_{A,B} &=& |A \cap B| \,, \\
\nu_{A,B} &=& |(A \cup B)^c| \,,
\end{eqnarray*}
then this family of similarity measures for sets can be expressed as:
\begin{eqnarray}
\label{eq:simfamily}
S(A,B) = \frac{t \Delta_{A,B} + u \delta_{A,B} + v \nu_{A,B}}{t' \Delta_{A,B} + u \delta_{A,B} + v \nu_{A,B}} \,,
\end{eqnarray}
with $t$, $t'$, $u$ and $v$ four parameters.
This family of similarity measures includes many well-known similarity measures for sets, such as the Jaccard coefficient \cite{Jaccard1908}, the simple matching coefficient \cite{Sokal1958} and the Dice coefficient \cite{Dice1945}.

Three members of this family are investigated in our experiments. The first one is the Jaccard coefficient, corresponding to $(t,t',u,v) = (0,1,1,0)$. The Jaccard coefficient is known to be $T_{\bf L}$-transitive. The second member that we investigate was originally proposed by \cite{Sokal1963}. It corresponds to $(t,t',u,v) = (0,1,2,2)$ and it does not satisfy $T_{\bf L}$-transitivity, which is considered as a very weak transitivity condition. Conversely, the third member that we analyse has rather strong transitivity properties. It is given by $(t,t',u,v) = (1,2,1,1)$ and it satisfies $T_{\bf P}$-transitivity.

\begin{figure}
\begin{center}
\includegraphics[scale=0.55]{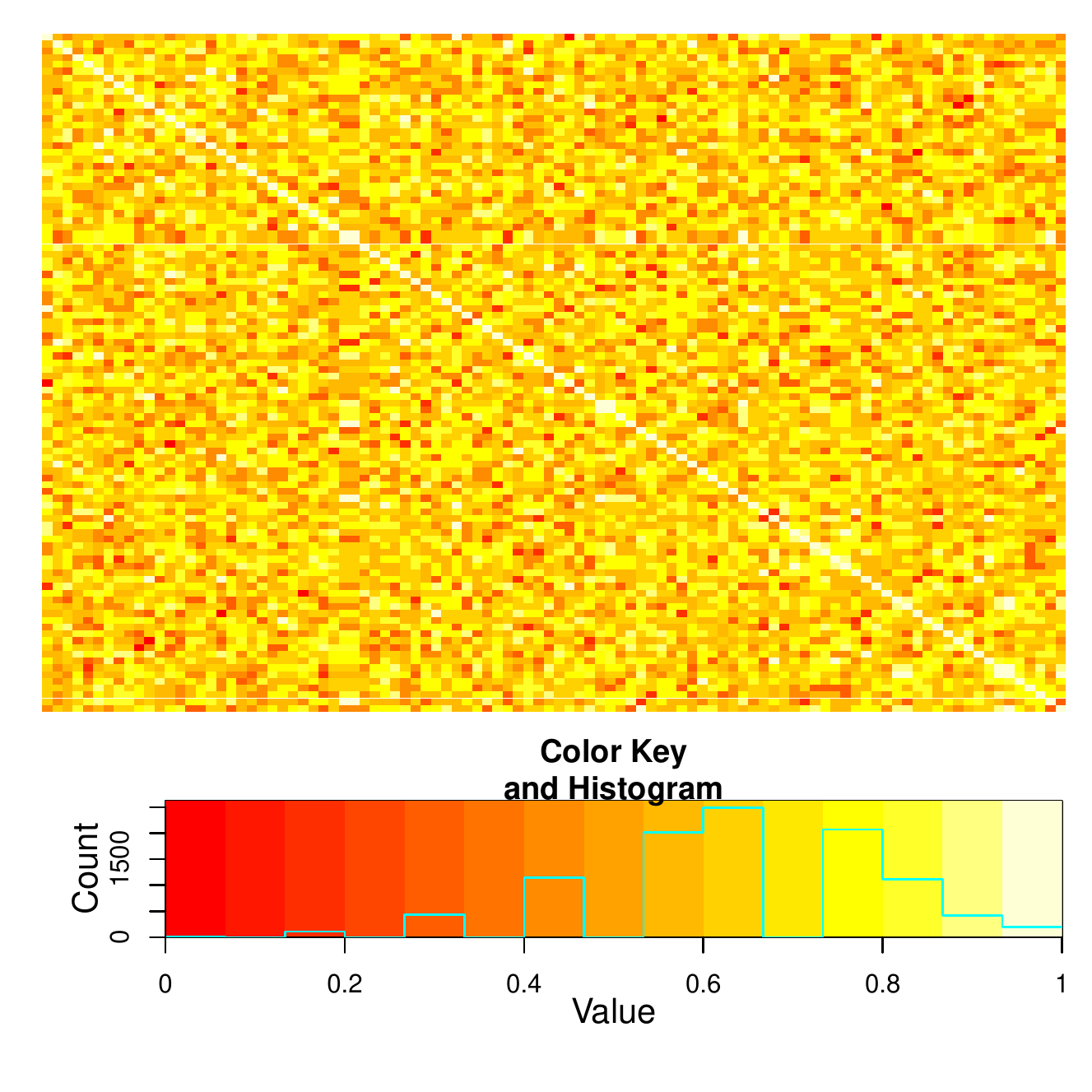}\\
\includegraphics[scale=0.55]{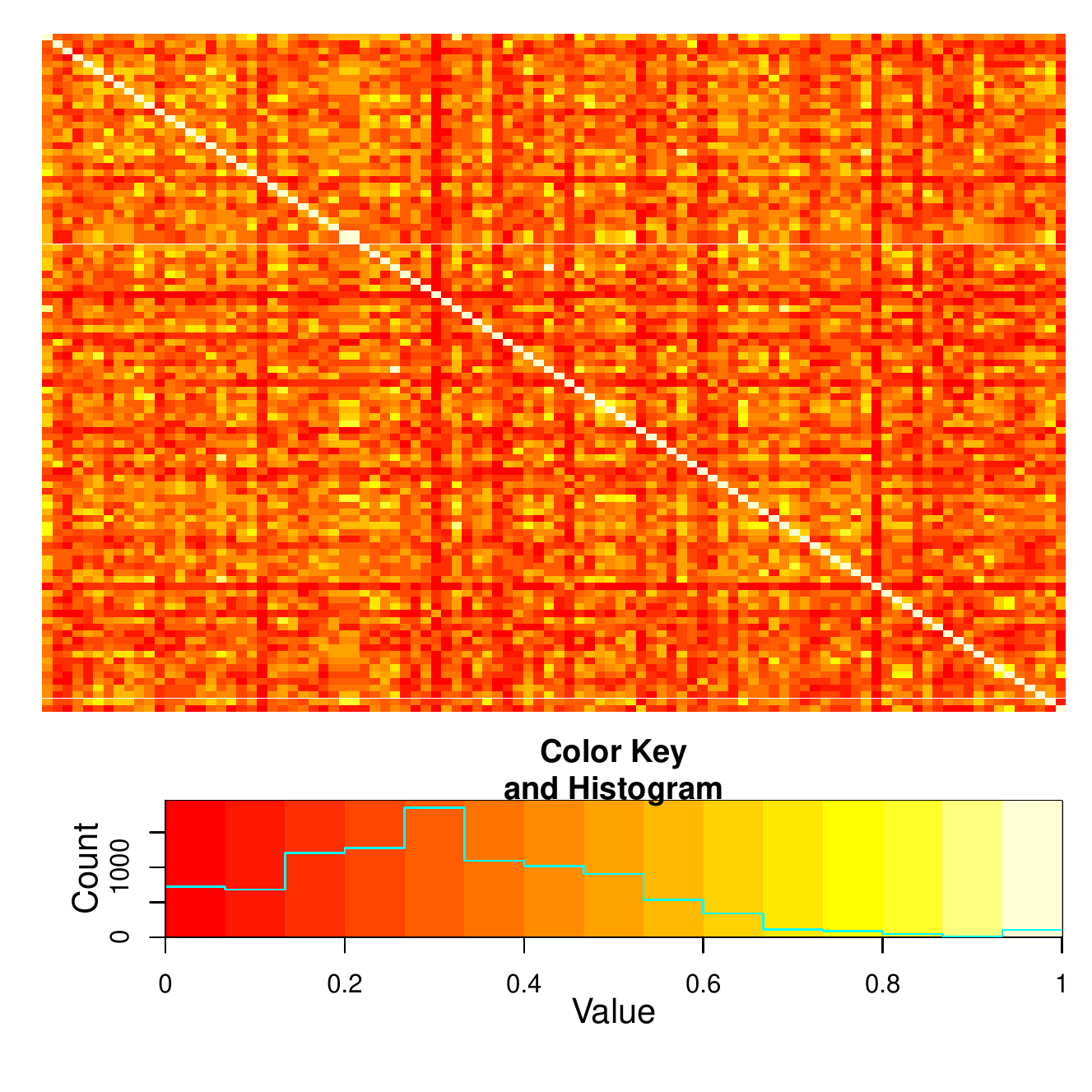}\\
\includegraphics[scale=0.55]{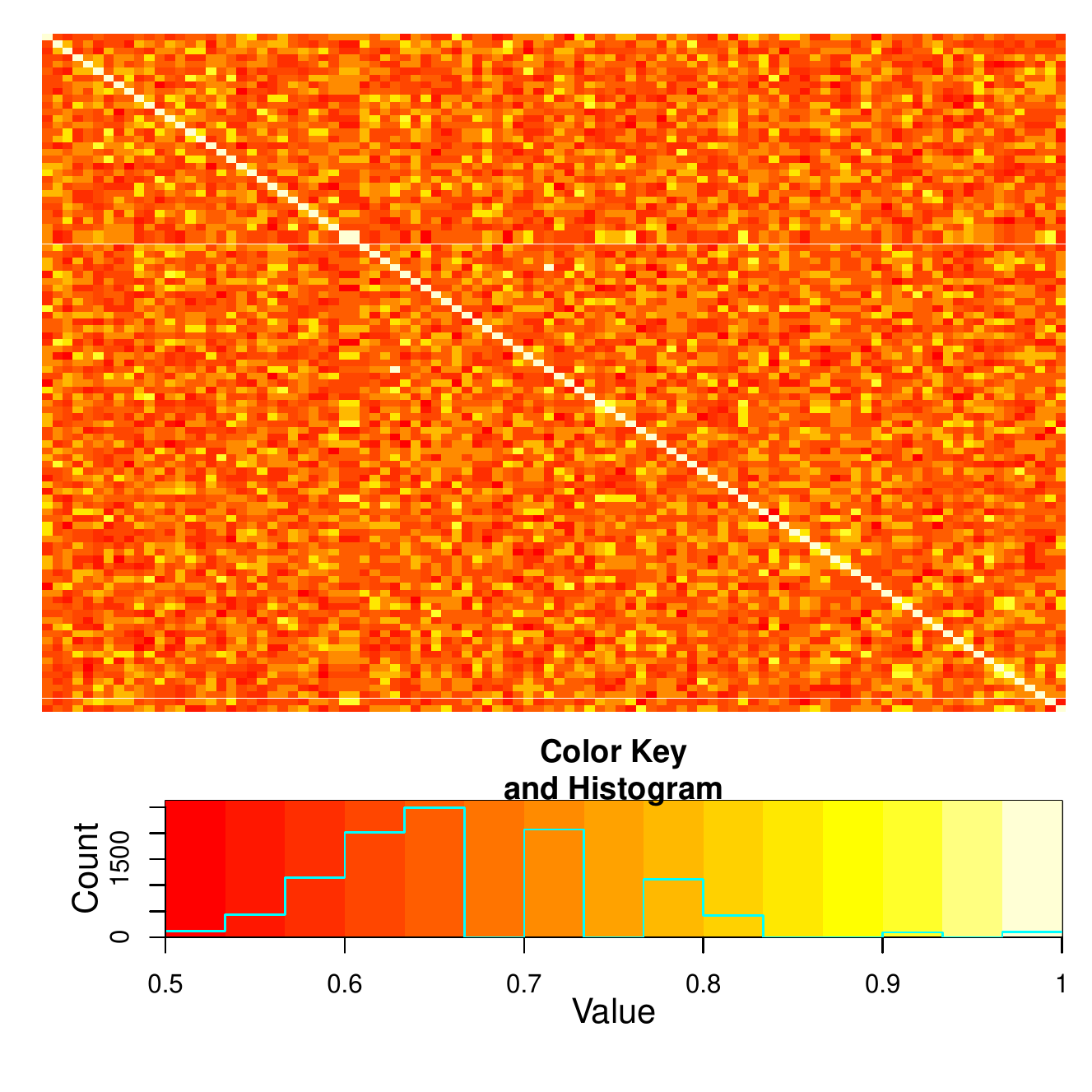}
\end{center}
\caption{The distribution of similarity scores obtained on a 100 by 100 matrix for all three members of the family. From top to bottom: $(t,t',u,v) = (0,1,2,2)$, $(t,t',u,v) = (0,1,1,0)$ and $(t,t',u,v) = (1,2,1,1)$.}
\label{fig:heatmaps}
\end{figure}

Features and labels for all three members are generated as follows. First we generate 20-dimensional feature vectors consisting of statistically independent features that follow a Bernoulli distribution with $\pi=0.5$. Subsequently, the above-mentioned similarity measures are computed for each pair of features, resulting in a deterministic mapping between features and labels. Finally, to introduce some noise in the problem setting, 10\% of the features are swapped in a last step from a zero to a one or vice versa. Figure~\ref{fig:heatmaps} illustrates the distribution of the obtained similarity scores for a 100 $\times$ 100 matrix.

In the experiments, we always generate three data sets, a training set for building the model, a validation set for hyperparameter selection, and a test set for performance evaluation. We perform two kinds of experiments. In the first experiment, we have a single set of $100$ nodes. $500$ node pairs are randomly sampled without replacement to the training, validation and test sets. Thus, the learning problem here is, given a subset of the relation values for a fixed set of nodes, to learn to predict missing relation values. This setup allows us to test also the Cartesian kernel, which is unable to generalize to completely new pairs of nodes. In the second experiment, we generate three separate sets of $100$ nodes for the training, validation and test sets, and sample from each of these $500$ edges. This experiment allows us to test the generalization capability of the learned models with respect to new couples of nodes (i.e., previously unseen nodes). Here, the Cartesian kernel is not applicable, and thus not included in the experiment. The experiments are repeated 100 times, the presented results are means over the repetitions. For statistical significance testing, we use the paired Wilcoxon-signed-rank test with significance level $0.05$. All pairs of kernels are compared, and the conservative Bonferroni correction is applied to take into account multiple hypothesis testing, meaning that the required p-value is divided by the number of comparisons. The Gaussian RBF kernel was considered at the node level. The used performance measure is the mean squared error (MSE). For training RLS we solve the corresponding system of linear equations using matrix factorization, by considering an explicit regularization parameter. A grid search is conducted to select the width of the Gaussian RBF kernel and the regularization parameter of the RLS algorithm. Both parameters are selected from the range $2^{-20}, \ldots, 2^1$.

\begin{table}
\centering
\begin{tabular}{lrccccccc}
\hline
Setting & $(t,t',u,v)$ & MPRED & $K_{\otimes}^{\Phi}$ & $K_{\otimes S}^{\Phi}$ & $K_{\rm MLPK}^{\Phi}$ & $K_{C}^{\Phi}$ & $K_{C S}^{\Phi}$  \\
\hline
\hline
Intransitive & (0,1,2,2) & 0.01038 & 0.00908 & 0.00773 & 0.00768 & 0.00989 & 0.00924\\
$T_{\bf L}$-transitive & (0,1,1,0) & 0.01514 & 0.00962 & 0.00781 & 0.00805 & 0.01155 & 0.00941\\
$T_{\bf P}$-transitive & (1,2,1,1) & 0.00259 & 0.00227 & 0.00192 & 0.00188 & 0.00248 & 0.00231\\
\hline
\end{tabular}
\centering
\caption{The predictive performance on test data for the different types of relations and kernels. In this experiment, the task is to predict relation values for unknown edges in a partially observed relational graph. The performance measure is the mean squared error.}
\label{table:artificial1}
\end{table}

\begin{table}
\centering
\begin{tabular}{lrccccc}
\hline
Setting & $(t,t',u,v)$ & MPRED & $K_{\otimes}^{\Phi}$ & $K_{\otimes S}^{\Phi}$ & $K_{\rm MLPK}^{\Phi}$  \\
\hline
\hline
Intransitive & (0,1,2,2) & 0.01032 & 0.00995 & 0.00936 & 0.00971\\
$T_{\bf L}$-transitive & (0,1,1,0) & 0.01515 & 0.01236 & 0.01166 & 0.01453\\
$T_{\bf P}$-transitive & (1,2,1,1) & 0.00259 & 0.00251 & 0.00236 & 0.00242\\
\hline
\end{tabular}
\centering
\caption{The predictive performance on test data for the different types of relations and kernels. In this experiment, the task is to predict relation values for a completely new set of nodes. The performance measure is the mean squared error.}
\label{table:artificial2}
\end{table}


The results for the experiments are presented in Tables~\ref{table:artificial1} and~\ref{table:artificial2}. In both cases all the kernels outperform the mean as prediction, meaning that they are able to model the underlying relations. For all the learning methods, the error is lower in the first experiment than in the second one, demonstrating that it is easier to predict relations between known nodes, than to generalize to a new set of nodes.
Enforcing symmetry is clearly beneficial, as the symmetric Kronecker product pairwise kernel always outperforms the standard Kronecker product pairwise kernel, and the symmetric Cartesian kernel always outperforms the standard one. Comparing the Kronecker and Cartesian kernels, the Kronecker one leads to clearly lower error rates. With the exception of the $T_{\bf L}$-transitive case in the second experiment, MLPK turns out to be highly successful in modeling the relations, probably due to enforcing symmetry of the learned relation. In the first experiment, all the differences are statistically significant, apart from the difference between the symmetric Kronecker product pairwise kernel and MLPK for the intransitive case. In the second experiment, all the differences are statistically significant. We can conclude that including prior knowledge about symmetry really helps boosting the predictive performance in this problem.

\subsection{Learning the similarity between documents}
In the second experiment, we compare the ordinary and symmetric Kronecker pairwise kernels on a real-world data set based on newsgroups documents\footnote{Available at: \url{http://people.csail.mit.edu/jrennie/20Newsgroups/}}. The data is sampled from 4 newsgroups: rec.autos, rec.sport.baseball, comp.sys.ibm.pc.hardware and comp.windows.x. The aim is to learn to predict the similarity of two documents as measured by the number of common words they share. The node features correspond to the number of occurrences of a word in a document. Unlike the previous experiment, the feature representation is very high-dimensional and sparse, as there are more than $50000$ possible features, the majority of which are zero for any given document. First, we sample separate training, validation and test sets each consisting of $1000$ nodes. Second, we sample edges connecting the nodes in the training and validation set using exponentially growing sample sizes to measure the effect of sample size on the differences between the kernels. The sample size grid is $[ 100, 200, 400, \ldots ,102400 ]$. Again, we sample only edges with different starting and end nodes. When computing the test performance, we consider all the edges in the test set, except those starting and ending at the same node.  The linear kernel is used at the node level.  We train the RLS algorithm using conjugate gradient optimization with early stopping \cite{Pahikkala2010conditional}, optimization is terminated once the MSE on the validation set has failed to decrease for 10 consecutive iterations.  Since we rely on the regularizing effect of early stopping, a separate regularization parameter is not needed in this experiment. We do not include other types of kernels than the Kronecker product pairwise kernels in the experiment. To the best of our knowledge, no algorithms that scale to the considered experiment size exist for the other kernel functions. Hence, this experiment mainly aims to illustrate the computational advantages of the Kronecker product pairwise kernel.
The mean as prediction achieves an MSE around $145$ on this dataset.

\begin{figure}
\begin{center}
\includegraphics[width=\linewidth]{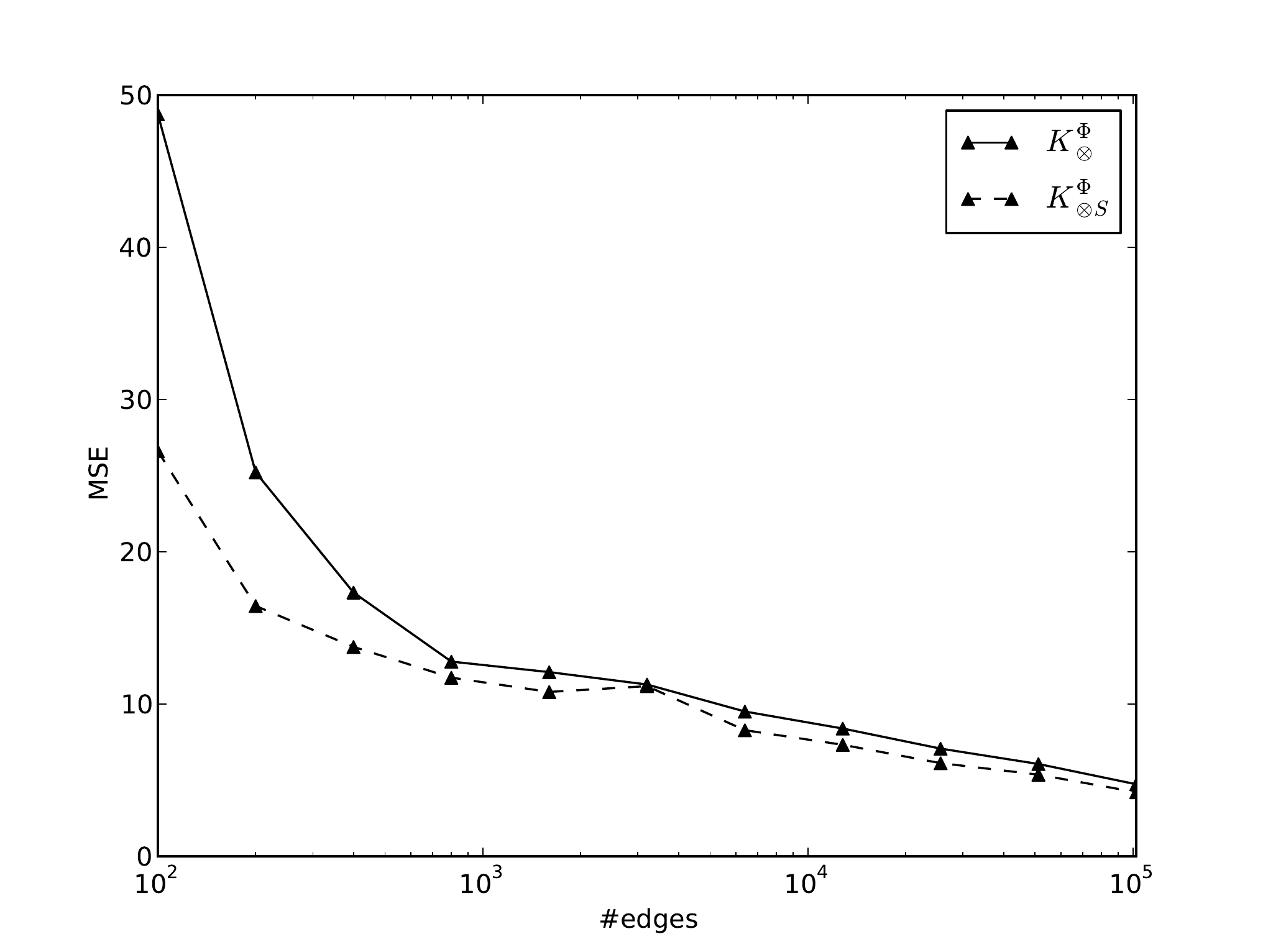}\\
\end{center}
\caption{The comparison of the ordinary Kronecker product pairwise kernel $K_{\otimes}^{\Phi}$ and the symmetric Kronecker product pairwise kernel $K_{\otimes S}^{\Phi}$ on the Newsgroups dataset. The mean squared error is shown as a function of the training set size.}
\label{fig:newsgroups}
\end{figure}

The results are presented in Figure~\ref{fig:newsgroups}. Even for $100$ pairs the errors are for both kernels much lower than the results for the mean as prediction, showing that the RLS algorithm succeeds with both kernels in learning the underlying relation. Increasing the training set size leads to a decrease in test error. Using the prior knowledge about the symmetry of the learned relation is clearly helpful. The symmetric kernel achieves for all sample sizes a lower error than the ordinary Kronecker product pairwise kernel and the largest differences are observed for the smallest sample sizes. For $100$ training instances, the error is almost halved by enforcing symmetry.

\subsection{Competition between species}
In this final experiment we evaluate the performance of the ordinary and reciprocal Kronecker pairwise kernels and the metric learning pairwise kernel on simulated data from an ecological model. The setup is based on the one described in \cite{allesina2011competitive}. This model provides an elegant explanation for the coexistence of multiple species in the same habitat, a problem that has puzzled ecologists for decades \cite{hutchinson1961paradox}.

Imagine $n$ species sharing a habitat and struggling for their share of the resources. One species can dominate another species based on $k$ so-called limiting factors. A limiting factor defines an attribute that can give a fitness advantage, for example in plants, such as the ability to photosynthesize, the ability to draw minerals from the soil, resistance to diseases, etc. Each species can score better or worse on each of its $k$ limiting factors. The degree to which one species can dominate a competitor is relative to the number of limiting factors for which it is superior. All possible interactions can thus be represented in a tournament. In this framework relations are reciprocal and often intransitive.

For this simulation 400 species were simulated with 10 limiting factors. The value of each limiting factor is for each species drawn from a random uniform distribution between 0 and 1. Thus, any species $v$ can be represented by a vector $\mathbf{f}$ of length $k$ with the limiting factors as elements. The probability that a species $v$ dominates species $v'$ can easily be calculated:
\begin{equation}
Q(v,v')=\frac{1}{k}\sum_{i=1}^{k} H(f_i-f_i'), \label{lf}
\end{equation}
where $H(x)$ is the Heaviside step function.

Of the 400 species, 200, 100 and 100 were used for generating training, validation and testing data. For each subset, the complete tournament matrix was determined using (\ref{lf}). From those matrices 1200 interactions were sampled for training, 600 for model validation and 600 for testing. No combination of species was used more than once.
Using the limiting factors as features, we try to regress the probability that one species dominates another one using the ordinary and reciprocal Kronecker product pairwise kernels and the metric learning pairwise kernel. Again, the Gaussian kernel is applied as the node kernel. The validation set is used to determine the optimal regularization parameter and kernel width parameter from the grids $ 2^{-20}$, $2^{-19}$ $\ldots$, $2^4$ and $2^{-10}$, $2^{-9}$ $\ldots$, $2^1$. To obtain statistically significant results the setup is repeated 100 times.\\

\begin{table}[h]
\begin{center}
\caption{The predictive performance on test data for the different types of  kernels. The performance measure is the mean squared error.}
\label{ecores}
\begin{tabular}{lccccc}
\hline
Kernel & MPRED & $K_{\otimes}^{\Phi}$ & $K_{\otimes R}^{\Phi}$ & $K_{\rm MLPK}^{\Phi}$ \\
\hline
\hline
MSE & 0.02795 & 0.01082 & 0.01067 & 0.02877 \\
\hline
\end{tabular}
\end{center}
\end{table}


The results are shown in Table~\ref{ecores}. The Wilcoxon-signed-rank test with significance level 0.05 is used for significance testing, and a conservative Bonferroni correction is applied for multiple hypothesis testing. All differences are statistically significant.

The metric learning pairwise kernel gives rise to worse predictions than the mean as prediction. This is not surprising, as the MLPK cannot learn reciprocal relations. The ordinary Kronecker product pairwise kernel performs good and the reciprocal Kronecker product pairwise kernel performs even better. All the differences are statistically significant. The results show that using the information on the types of relations to be learned can boost the accuracy of the predictions.

\section{Conclusion}
A general kernel-based framework for learning various types of graded relations was presented in this article. This framework extends existing approaches for learning relations, because it can handle crisp and graded relations. A Kronecker product feature mapping was proposed for combining the features of pairs of objects that constitute a relation (edge level in a graph), and it was shown that this mapping leads to a class of universal approximators, if an appropriate kernel is chosen on the object level (node level in a graph).

In addition, we clarified that domain knowledge about the relation to be learned can be easily incorporated in our framework, such as reciprocity and symmetry properties. Experimental results on synthetic and real-world data clearly demonstrate that this domain knowledge really helps in improving the generalization performance. Moreover, important links with recent developments in fuzzy set theory and decision theory can be established, by looking at transitivity properties of relations.


\section*{Acknowledgments}
W.W. is supported as a postdoc by the Research Foundation of Flanders (FWO Vlaanderen) and T.P. by the Academy of Finland (grant 134020).

\section*{Appendix}
\subsection{Formal definitions}
\begin{definition}
The Kronecker product of two matrices $\anymatrix$ and $\othermatrix$ is defined as
\begin{eqnarray*}
\anymatrix\otimes\othermatrix=\left(
\begin{array}{ccc}
{\anymatrix}_{1,1}\othermatrix&\cdots&\anymatrix_{1,n}\othermatrix\\
\vdots&\ddots&\vdots\\
{\anymatrix}_{m,1}\othermatrix&\cdots&\anymatrix_{m,n}\othermatrix
\end{array}
\right),
\end{eqnarray*}
\end{definition}

\begin{definition}[\cite{Steinwart2002consistency}]
A continuous kernel $\kernelf$ on a compact metric space $\mathcal{V}$ (i.e. $\mathcal{V}$ is closed and bounded) is called universal if the RKHS induced by $\kernelf$ is dense in $C(\mathcal{V})$, where $C(\mathcal{V})$ is the space of all continuous functions $f : \mathcal{V} \rightarrow \mathbb{R}$. That is, for every function $f\in C(\mathcal{V})$ and every $\epsilon > 0$, there exists a set of input points $\{\node_i \}_{i=1}^m \in \mathcal{V}$ and real numbers $\{\alpha_i\}_{i=1}^m$, with $m\in \mathbb{N}$, such that
\begin{equation*}
\max_{x\in \mathcal{V}}\left\{\left\arrowvert f(\node)-\sum_{i=1}^m\alpha_i\kernelf(\node_i,\node)\right\arrowvert\right\}\leq\epsilon.
\end{equation*}
Accordingly, the hypothesis space induced by the kernel $\kernelf$ can approximate any function in $C(\mathcal{V})$ arbitrarily well, and hence it has the universal approximating property.
\end{definition}
The following result is in the literature known as the Stone-Weierstra{\ss} theorem (see e.g \cite{rudin1991functional}):
\begin{theorem}[Stone-Weierstra{\ss}]\label{stoneweierstrass}
Let $\mathcal{V}$ be a compact metric space and let $C(\mathcal{V})$ be the set of real-valued continuous functions on $\mathcal{V}$. If $\mathcal{A}\subset C(\mathcal{V})$ is a subalgebra of $C(\mathcal{V})$, that is,
\begin{equation*}
\begin{array}{l}
\forall {f(\node),g(\node)\in\mathcal{A}}, {r\in\mathbb{R}}:
{f(\node)+rg(\node)\in\mathcal{A}, f(\node)g(\node)\in\mathcal{A}}
\end{array}
\end{equation*}
and $\mathcal{A}$ separates points in $\mathcal{V}$, that is,
\begin{equation*}
\forall \node,\node'\in\mathcal{V},\node \neq \node':\exists g\in\mathcal{A}:g(\node)\neq g(\node'),
\end{equation*}
and $\mathcal{A}$ does not vanish at any point in $\mathcal{V}$, that is,
\begin{equation*}
\forall \node \in\mathcal{V}:\exists g\in\mathcal{A}:g(\node)\neq 0,
\end{equation*}
then $\mathcal{A}$ is dense in $C(\mathcal{V})$.
\end{theorem}

\subsection{Proofs}

\begin{proof}({\bf Theorem~\ref{unikrontheorem}})
Let us define
\begin{equation}\label{funckron}
\begin{array}{l}
\mathcal{A}\otimes\mathcal{A}
=\left\{t\mid t(\node,\node')=g(\node)u(\node'),g,u\in \mathcal{A}\right\}
\end{array}
\end{equation}
for a compact metric space $\mathcal{V}$ and a set of functions $\mathcal{A}\subset C(\mathcal{V})$. We observe that the RKHS of the kernel $\kernelf_{\otimes}^\Phi$ can be written as $\mathcal{H}\otimes\mathcal{H}$, where $\mathcal{H}$ is the RKHS of the kernel $\kernelf^\phi$.

Let $\epsilon>0$ and let $t\in C(\nodeset)\otimes C(\nodeset)$ be an arbitrary function which can, according to (\ref{funckron}), be written  as $t(\node,\node')=g(\node)u(\node')$, where $g,u\in C(\nodeset)$. By definition of the universality property, $\mathcal{H}$ is dense in $C(\nodeset)$. Therefore, $\mathcal{H}$ contains functions $\overline{g},\overline{u}$ such that
\[
\max_{\node\in\nodeset}\left\{\left\arrowvert \overline{g}(\node)-g(\node)\right\arrowvert\right\}\leq \overline{\epsilon},\ \max_{\node\in\nodeset}\left\{\left\arrowvert \overline{u}(\node)-u(\node)\right\arrowvert\right\}\leq \overline{\epsilon} \,,
\]
where $\overline{\epsilon}$ is a constant for which it holds that
\[
\max_{\node,\node'\in\nodeset}\left\{\left\arrowvert \overline{\epsilon} \, \overline{g}(\node)\right\arrowvert+\left\arrowvert\overline{\epsilon} \,\overline{u}(\node')\right\arrowvert+\overline{\epsilon}^2\right\}\leq \epsilon \,.
\]
Note that, according to the extreme value theorem, the maximum exists due to the compactness of $\nodeset$ and the continuity of the functions $g$ and $u$. Now we have
\[
\begin{array}{l}
\displaystyle
\max_{\node,\node'\in\nodeset}\left\{\left\arrowvert t(\node,\node')-\overline{g}(\node)\overline{u}(\node')\right\arrowvert\right\}\\
\displaystyle
\leq\max_{\node,\node'\in\nodeset}\left\{\left\arrowvert t(\node,\node')-g(\node)u(\node')\right\arrowvert+\left\arrowvert \overline{\epsilon} \,\overline{g}(\node)\right\arrowvert+\left\arrowvert\overline{\epsilon}\,\overline{u}(\node')\right\arrowvert+\overline{\epsilon}^2\right\}\\
\displaystyle
=\max_{\node,\node'\in\nodeset}\left\{\left\arrowvert \overline{\epsilon} \,\overline{g}(\node)\right\arrowvert+\left\arrowvert\overline{\epsilon} \,\overline{u}(\node')\right\arrowvert+\overline{\epsilon}^2\right\}\\
\displaystyle\leq \epsilon,
\end{array}
\]
which confirms the density of $\mathcal{H}\otimes\mathcal{H}$ in $C(\nodeset)\otimes C(\nodeset)$.

According to Tychonoff's theorem, $\nodeset^2$ is compact if $\nodeset$ is compact. It is straightforward to see that $C(\nodeset)\otimes C(\nodeset)$ is a subalgebra of $C(\nodeset^2)$, it separates points in $\nodeset^2$, it vanishes at no point of $C(\nodeset^2)$, and it is therefore dense in $C(\nodeset^2)$ due to Theorem~\ref{stoneweierstrass}. Consequently, $\mathcal{H}\otimes\mathcal{H}$ is also dense in $C(\nodeset^2)$, and $\kernelf_{\otimes}^\Phi$ is a universal kernel on $\edgeset$.
\end{proof}

\begin{proof} ({\bf Theorem~\ref{antisymmetrictheorem}})
Let $\epsilon>0$ and $t\in R(\nodeset^2)$ be an arbitrary function. According to Theorem~\ref{unikrontheorem}, the RKHS of the kernel $\kernelf_{\otimes}^\Phi$ defined in (\ref{eq:tppk}) is dense in $C(\nodeset^2)$. Therefore, we can select a set of edges and real numbers $\{\alpha_i\}_{i=1}^m$, such that the function
\begin{equation*}
u(\node,\node')=\sum_{i=1}^m\alpha_i\kernelf^{\phi}(\node,\node_i) \kernelf^{\phi}(\node',\node_i')
\end{equation*}
belonging to the RKHS of the kernel (\ref{eq:tppk}) fulfills
\begin{equation}\label{tempapproxone}
\max_{(\node,\node')\in \nodeset^2}\left\{\left\arrowvert t(\node,\node')-4u(\node,\node')\right\arrowvert\right\}\leq\frac{1}{2}\epsilon \,.
\end{equation}
We observe that, because $t(\node,\node')=-t(\node',\node)$, the function $u$ also fulfills
\begin{equation*}
\max_{(\node,\node')\in \nodeset^2}\left\{\left\arrowvert t(\node,\node')+4u(\node',\node)\right\arrowvert\right\}\leq\frac{1}{2}\epsilon
\end{equation*}
and hence
\begin{equation}\label{tempapproxthree}
\max_{(\node,\node')\in \nodeset^2}\left\{\left\arrowvert 4u(\node,\node')+4u(\node',\node)\right\arrowvert\right\}\leq\epsilon \,.
\end{equation}
Let
\[
\gamma(\node,\node')=2u(\node,\node')+2u(\node',\node) \,.
\]
Due to (\ref{tempapproxthree}), we have
\begin{equation}\label{gammaineq}
\arrowvert\gamma(\node,\node')\arrowvert\leq\frac{1}{2}\epsilon,\phantom{\qed}\forall(\node,\node')\in\nodeset^2 \,.
\end{equation}
Now, let us consider the function $\hypothesis(\node,\node')=$
\begin{equation*}
\sum_{i=1}^m\alpha_i 2\left(\kernelf^{\phi}(\node,\node_i) \kernelf^{\phi}(\node',\node_i')-\kernelf^{\phi}(\node',\node_i) \kernelf^{\phi}(\node,\node_i')\right) \,,
\end{equation*}
which is obtained from $u$ by replacing kernel (\ref{eq:tppk}) with kernel (\ref{eq:recedgekernel}). We observe that
{\setlength\arraycolsep{2pt}
\begin{eqnarray}
\hypothesis(\node,\node')&=&2u(\node,\node')-2u(\node',\node)\nonumber\\
&=&4u(\node,\node')-\gamma(\node,\node')\label{fgammaeq}.
\end{eqnarray}
}
By combining (\ref{tempapproxone}), (\ref{gammaineq}) and (\ref{fgammaeq}), we observe that the function $\hypothesis$ fulfills (\ref{recclaim}).

\end{proof}

\begin{proof} ({\bf Proposition~\ref{thm:mlpk}})
The model that we consider can be written as:
$$Q(\node,\node') = \nabla \big((\phi(\node) - \phi(\node'))^T \bm{W} (\phi(\node) - \phi(\node')) \big) \,.$$
The connection with (\ref{eq:rrsym}) then immediately follows by decomposing $\bm{W}$ as $\bm{W} = \bm{U}^T \bm{U}$ with $\bm{U}$ an arbitrary matrix. The specific case of $z=1$ is obtained when $\bm{U}$ can be written as a single-row matrix.
\end{proof}

\bibliography{referenties,myBibliography}

\begin{thebibliography}{10}

\bibitem{allesina2011competitive}
Stefano Allesina and Jonathan~M. Levine.
\newblock A competitive network theory of species diversity.
\newblock {\em Proceedings of the National Academy of Sciences},
  108:5638--5642, 2011.

\bibitem{Belkin2006}
M.~Belkin, P.~Niyogi, and V.~Sindhwani.
\newblock Manifold regularization: a geometric framework for learning from
  labeled and unlabeled examples.
\newblock {\em Journal of Machine Learning Research}, 7:2399--2434, 2006.

\bibitem{Ben-Hur2005}
A.~{Ben-Hur} and W.~Noble.
\newblock Kernel methods for predicting protein-protein interactions.
\newblock {\em Bioinformatics}, 21 Suppl 1:38--46, 2005.

\bibitem{Billot1995}
A.~Billot.
\newblock An existence theorem for fuzzy utility functions: A new elementary
  proof.
\newblock {\em Fuzzy Sets and Systems}, 74:271--276, 1995.

\bibitem{Boddy2000}
L.~Boddy.
\newblock Interspecific combative interactions between wood-decaying
  basidiomycetes.
\newblock {\em {F}{E}{M}{S} Microbiology Ecology}, 31:185--194, 2000.

\bibitem{Bodenhofer2007}
U.~Bodenhofer, B.~{De Baets}, and J.~Fodor.
\newblock A compendium of fuzzy weak orders.
\newblock {\em Fuzzy Sets and Systems}, 158:811--829, 2007.

\bibitem{Bowling2006}
M.~Bowling, J.~F{\"u}rnkranz, T.~Graepel, and R.~Musick.
\newblock Machine learning and games.
\newblock {\em Machine Learning}, 63(3):211--215, 2006.

\bibitem{czaran2002chemical}
T.~Cz{\'a}r{\'a}n, R.~Hoekstra, and L.~Pagie.
\newblock Chemical warfare between microbes promotes biodiversity.
\newblock {\em Proceedings of the National Academy of Sciences},
  99(2):786--790, 2002.

\bibitem{DeBaets2005}
B.~{De Baets} and H.~{De Meyer}.
\newblock Transitivity frameworks for reciprocal relations: cycle-transitivity
  versus {$F$}{$G$}-transitivity.
\newblock {\em Fuzzy Sets and Systems}, 152:249--270, 2005.

\bibitem{DeBaets2006}
B.~{De Baets}, H.~{De Meyer}, B.~{De Schuymer}, and S.~Jenei.
\newblock Cyclic evaluation of transitivity of reciprocal relations.
\newblock {\em Social Choice and Welfare}, 26:217--238, 2006.

\bibitem{DeBaets2001}
B.~{De Baets}, H.~{De Meyer}, and H.~Naessens.
\newblock A class of rational cardinality-based similarity measures.
\newblock {\em J. Comput. Appl. Math.}, 132:51--69, 2001.

\bibitem{DeBaets2002}
B.~{De Baets} and R.~Mesiar.
\newblock Metrics and {$T$}-equalities.
\newblock {\em Journal of Mathematical Analysis and Applications},
  267:531--547, 2002.

\bibitem{Deraedt2009}
L.~{De Raedt}.
\newblock {\em Logical and Relational Learning}.
\newblock Springer, 2009.

\bibitem{DeSchuymer2003}
B.~{De Schuymer}, H.~{De Meyer}, B.~{De Baets}, and S.~Jenei.
\newblock On the cycle-transitivity of the dice model.
\newblock {\em Theory and Decision}, 54:261--285, 2003.

\bibitem{Diaz2007}
S.~Diaz, S.~Montes, and B.~{De Baets}.
\newblock Transitivity bounds in additive fuzzy preference structures.
\newblock {\em {I}{E}{E}{E} Transactions on Fuzzy Systems}, 15:275--286, 2007.

\bibitem{Dice1945}
L.~Dice.
\newblock Measures of the amount of ecologic associations between species.
\newblock {\em Ecology}, 26:297--302, 1945.

\bibitem{Doignon1986}
{J.-P}. Doignon, B.~Monjardet, M.~Roubens, and {Ph.} Vincke.
\newblock Biorder families, valued relations and preference modelling.
\newblock {\em Journal of Mathematical Psychology}, 30:435--480, 1986.

\bibitem{Fishburn1991}
P.~Fishburn.
\newblock Nontransitive preferences in decision theory.
\newblock {\em Journal of Risk and Uncertainty}, 4:113--134, 1991.

\bibitem{Fisher2008}
L.~Fisher.
\newblock {\em Rock, Paper, Scissors: Game Theory in Everyday Life}.
\newblock Basic Books, 2008.

\bibitem{Fono2007}
L.~Fono and N.~Andjiga.
\newblock Utility function of fuzzy preferences on a countable set under
  max-*-transitivity.
\newblock {\em Social Choice and Welfare}, 28:667--683, 2007.

\bibitem{Geurts2007}
P.~Geurts, N.~Touleimat, M.~Dutreix, and F.~{d'Alch{\'e}-Buc}.
\newblock Inferring biological networks with output kernel trees.
\newblock {\em BMC Bioinformatics}, 8(2):S4, 2007.

\bibitem{Gower1986}
J.~Gower and P.~Legendre.
\newblock Metric and {E}uclidean properties of dissimilarity coefficients.
\newblock {\em Journal of Classification}, 3:5--48, 1986.

\bibitem{Herbrich2000}
R.~Herbrich, T.~Graepel, and K.~Obermayer.
\newblock Large margin rank boundaries for ordinal regression.
\newblock In A.~Smola, P.~Bartlett, B.~Sch\"olkopf, and D.~Schuurmans, editors,
  {\em Advances in Large Margin Classifiers}, pages 115--132. MIT Press, 2000.

\bibitem{Hue2010}
M.~Hue and {J.-P.} Vert.
\newblock On learning with kernels for unordered pairs.
\newblock In {\em Proceedings of the 27th International Conference on Machine
  Learning, p.463-470, 2010}, 2010.

\bibitem{Hullermeier2010a}
E.~H{\"ullermeier} and J.~F{\"u}rnkranz.
\newblock {\em Preference Learning}.
\newblock Springer, 2010.

\bibitem{hutchinson1961paradox}
G.~E. Hutchinson.
\newblock The paradox of the plankton.
\newblock {\em The American Naturalist}, 95(882):137--145, 1961.

\bibitem{Jaccard1908}
P.~Jaccard.
\newblock Nouvelle recherches sur la distribution florale.
\newblock {\em Bulletin de la Soci{\'e}t{\'e}e Vaudoise de Sciences
  Naturelles}, 44:223--270, 1908.

\bibitem{Jacob2008}
L.~Jacob and {J.-P.} Vert.
\newblock Protein-ligand interaction prediction: an improved chemogenomics
  approach", bioinformatics, 24(19):2149-2156, 2008.
\newblock {\em Bioinformatics}, 241:2149--2156, 2008.

\bibitem{Jakel2008}
F.~J{\"a}kel, B.~Sch{\"o}lkopf, and F.~Wichmann.
\newblock Similarity, kernels, and the triangle inequality.
\newblock {\em Journal of Mathematical Psychology}, 52(2):297--303, 2008.

\bibitem{karolyi2005rps}
G.~K{\'a}rolyi, Z.~Neufeld, and I.~Scheuring.
\newblock Rock-scissors-paper game in a chaotic flow: The effect of dispersion
  on the cyclic competition of microorganisms.
\newblock {\em Journal of Theoretical Biology}, 236(1):12--20, 2005.

\bibitem{Kashima2009}
H.~Kashima, S.~Oyama, Y.~Yamanishi, and K.~Tsuda.
\newblock On pairwise kernels: An efficient alternative and generalization
  analysis.
\newblock In Thanaruk Theeramunkong, Boonserm Kijsirikul, Nick Cercone, and
  Tu~Bao Ho, editors, {\em PAKDD}, volume 5476 of {\em Lecture Notes in
  Computer Science}, pages 1030--1037. Springer, 2009.

\bibitem{Kerr2002}
B.~Kerr, M.~Riley, M.~Feldman, and B.~Bohannan.
\newblock Local dispersal promotes biodiversity in a real-life game of rock
  paper scissors.
\newblock {\em Nature}, 418:171--174, 2002.

\bibitem{Kirkup2004}
B.~Kirkup and M.~Riley.
\newblock Antibiotic-mediated antagonism leads to a bacterial game of
  rock-paper-scissors in vivo.
\newblock {\em Nature}, 428:412--414, 2004.

\bibitem{Koppen1995}
M.~Koppen.
\newblock Random utility representation of binary choice probabilities:
  Critical graphs yielding critical necessary conditions.
\newblock {\em Journal of Mathematical Psychology}, 39:21--39, 1995.

\bibitem{Lawrence2009}
N.~Lawrence and R.~Urtasan.
\newblock Nonlinear matrix factorization with gaussian processes.
\newblock In {\em Proceedings of the International Conference on Machine
  Learning}, pages 601--608, 2009.

\bibitem{Luce1965}
R.~Luce and P.~Suppes.
\newblock {\em Handbook of Mathematical Psychology}, chapter Preference,
  Utility and Subjective Probability, pages 249--410.
\newblock Wiley, 1965.

\bibitem{Menon2010}
A.~Menon and C.~Elkan.
\newblock Predicting labels for dyadic data.
\newblock {\em Data Mining and Knowledge Discovery}, 21:327--343, 2010.

\bibitem{Miller2009}
K.~Miller, T~Griffiths, and M.~Jordan.
\newblock Nonparametric latent feature models for link prediction.
\newblock {\em Advances in Neural Processing Systems}, 22:1276--1284, 2009.

\bibitem{Moser2006}
B.~Moser.
\newblock On representing and generating kernels by fuzzy equivalence
  relations.
\newblock {\em Journal of Machine Learning Research}, 7:2603--2620, 2006.

\bibitem{nowak2002}
M.~Nowak.
\newblock Biodiversity: Bacterial game dynamics.
\newblock {\em Nature}, 418:138--139, 2002.

\bibitem{pahikkala2009preferences}
T.~Pahikkala, E.~Tsivtsivadze, A.~Airola, J.~J{\"a}rvinen, and J.~Boberg.
\newblock An efficient algorithm for learning to rank from preference graphs.
\newblock {\em Machine Learning}, 75(1):129--165, 2009.

\bibitem{Pahikkala2010conditional}
T.~Pahikkala, W.~Waegeman, A.~Airola, T.~Salakoski, and B.~{De Baets}.
\newblock Conditional ranking on relational data.
\newblock In J.~Balcázar, F.~Bonchi, A.~Gionis, and M.~Sebag, editors, {\em
  Proceedings of the European Conference on Machine Learning}, volume 6322 of
  {\em Lecture Notes in Computer Science}, pages 499--514. Springer Berlin /
  Heidelberg, 2010.

\bibitem{Pahikkala2010}
T.~Pahikkala, W.~Waegeman, E.~Tsivtsivadze, T.~Salakoski, and B.~{De Baets}.
\newblock Learning intransitive reciprocal relations with kernel methods.
\newblock {\em European Journal of Operational Research}, 206:676--685, 2010.

\bibitem{reichenbach2007}
T.~Reichenbach, M.~Mobilia, and E.~Frey.
\newblock Mobility promotes and jeopardizes biodiversity in rock-paper-scissors
  games.
\newblock {\em Nature}, 448:1046--1049, 2007.

\bibitem{rudin1991functional}
Walter Rudin.
\newblock {\em Functional Analysis}.
\newblock International Series in Pure and Applied Mathematics. McGraw-Hill
  Inc., New York, second edition, 1991.

\bibitem{Saunders1998}
C.~Saunders, A.~Gammerman, and V.~Vovk.
\newblock Ridge regression learning algorithm in dual variables.
\newblock In {\em Proceedings of the International Conference on Machine
  Learning}, pages 515--521, 1998.

\bibitem{Scholkopf2002}
B.~Sch\"olkopf and A.~Smola.
\newblock {\em Learning with Kernels, Support Vector Machines, Regularisation,
  Optimization and Beyond}.
\newblock The MIT Press, 2002.

\bibitem{Shawetaylor2004}
J.~{Shawe-Taylor} and N.~Cristianini.
\newblock {\em Kernel Methods for Pattern Analysis}.
\newblock Cambridge University Press, 2004.

\bibitem{Sinervo1996}
S.~Sinervo and C.~Lively.
\newblock The rock-paper-scissors game and the evolution of alternative mate
  strategies.
\newblock {\em Nature}, 340:240--246, 1996.

\bibitem{Sokal1958}
R.~Sokal and C.~Michener.
\newblock A statistical method for evaluating systematic relationships.
\newblock {\em Univ. of Kansas Science Bulletin}, 38:1409--1438, 1958.

\bibitem{Sokal1963}
R.~Sokal and P.~Sneath.
\newblock {\em Principles of Numerical Taxonomy}.
\newblock W. H. Freeman, 1963.

\bibitem{Srebro2005}
N.~Srebro, J.~Rennie, and T.~Jaakkola.
\newblock Maxximum margin matrix factorization.
\newblock {\em Advances in Neural Processing Systems}, 17, 2005.

\bibitem{Steinwart2002consistency}
I.~Steinwart.
\newblock On the influence of the kernel on the consistency of support vector
  machines.
\newblock {\em Journal of Machine Learning Research}, 2:67--93, 2002.

\bibitem{Suykens2002}
J.~Suykens, T.~{Van Gestel}, J.~{De Brabanter}, B.~{De Moor}, and
  J.~Vandewalle.
\newblock {\em Least Squares Support Vector Machines}.
\newblock World Scientific Pub. Co., Singapore, 2002.

\bibitem{Switalski2000}
Z.~Switalski.
\newblock Transitivity of fuzzy preference relations - an empirical study.
\newblock {\em Fuzzy Sets and Systems}, 118:503--508, 2000.

\bibitem{Switalski2003}
Z.~Switalski.
\newblock General transitivity conditions for fuzzy reciprocal preference
  matrices.
\newblock {\em Fuzzy Sets and Systems}, 137:85--100, 2003.

\bibitem{Taskar2004}
B.~Taskar, M.~Wong, P.~Abbeel, and D.~Koller.
\newblock Link prediction in relational data.
\newblock In {\em Advances in Neural Information Processing Systems}, 2004.

\bibitem{Tsochantaridis2005}
Y.~Tsochantaridis, T.~Joachims, T.~Hofmann, and Y.~Altun.
\newblock Large margin methods for structured and independent output variables.
\newblock {\em Journal of Machine Learning Research}, 6:1453--1484, 2005.

\bibitem{Tversky1998}
A.~Tversky.
\newblock {\em Preference, Belief and Similarity}.
\newblock {M}{I}{T} Press, 1998.

\bibitem{Vert2007}
{J.-P.} Vert, J.~Qiu, and W.~S. Noble.
\newblock A new pairwise kernel for biological network inference with support
  vector machines.
\newblock {\em BMC Bioinformatics}, 8 (Suppl 10):S8, 2007.

\bibitem{Vert2005}
J.-P. Vert and Y.~Yamanishi.
\newblock Supervised graph inference.
\newblock In {\em Advances in Neural Information Processing Systems},
  volume~17, 2005.

\bibitem{Vishwanathan2010}
S.~Vishwanathan, N.~Schraudolph, R.~Kondor, and K.~Borgwardt.
\newblock Graph kernels.
\newblock {\em Journal of Machine Learning Research}, 11:1201--1242, 2010.

\bibitem{Waite2001}
T.~Waite.
\newblock Intransitive preferences in hoarding gray jays (\textit{Perisoreus
  canadensis}).
\newblock {\em Journal of Behavioural Ecology and Sociobiology}, 50:116--121,
  2001.

\bibitem{Weston2004}
J.~Weston, A.~Eliseeff, D.~Zhou, C.~Leslie, and W.~Stafford Noble.
\newblock Protein ranking: from local to global structure in the protein
  similarity networks.
\newblock {\em Proceedings of the National Academy of Science}, 101:6559--6563,
  2004.

\bibitem{Weston2007}
J.~Weston, B.~Sch{\"o}lkopf, O.~Bousquet, T.~Mann, and W.~Noble.
\newblock {\em Predicting structured data}, chapter Joint kernel maps, pages
  67--83.
\newblock {M}{I}{T} Press, 2007.

\bibitem{Xing2002}
E.~Xing, A.~Ng, M.~Jordan, and S.~Russell.
\newblock Distance metric learning with application to clustering with side
  information.
\newblock In {\em Advances in Neural Information Processing Systems},
  volume~16, pages 521--528, 2002.

\bibitem{Yamanishi2004}
Y.~Yamanishi, {J.-P.} Vert, and M.~Kanehisa.
\newblock Protein network inference from multiple genomic data: a supervised
  approach.
\newblock {\em Bioinformatics}, 20:1363--1370, 2004.

\bibitem{Yang2009}
Y.~Yang, N.~Bansal, W.~Dakka, P.~Ipeirotis, N.~Koudas, and D.~Papadias.
\newblock Query by document.
\newblock In {\em Proceedings of the Second ACM International Conference on Web
  Search and Data Mining, Barcelona, Spain}, pages 34--43, 2009.

\bibitem{Zhang2003}
Z.~Zhang.
\newblock Learning metrics via discriminant kernels and multidimensional
  scaling: Toward expected {E}uclidean representation.
\newblock In {\em Proceedings of the Twentieth International Conference on
  Machine Learning, Washington D.C., USA}, pages 872--879, 2003.

\end{thebibliography}
\end{document}